\documentclass[preprint,11pt]{elsarticle}
\usepackage{amsmath,amssymb,graphicx,url}
\usepackage{caption,subcaption}
\usepackage{hyperref}
\usepackage{lineno}
\usepackage{multirow}
\usepackage{booktabs}
\usepackage{geometry}
\usepackage{arydshln}
\usepackage{placeins}
\usepackage{setspace}
\usepackage[table]{xcolor}

\journal{Journal}
\makeatletter
\def\adl@drawiv#1#2#3{%
        \hskip.5\tabcolsep
        \xleaders#3{#2.5\@tempdimb #1{1}#2.5\@tempdimb}%
                #2\z@ plus1fil minus1fil\relax
        \hskip.5\tabcolsep}
\newcommand{\cdashlinelr}[1]{%
  \noalign{\vskip\aboverulesep
           \global\let\@dashdrawstore\adl@draw
           \global\let\adl@draw\adl@drawiv}
  \cdashline{#1}
  \noalign{\global\let\adl@draw\@dashdrawstore
           \vskip\belowrulesep}}
\makeatother
\begin{document}
\onehalfspacing

\definecolor{nice_yellow}{HTML}{ffe49c}
\definecolor{nice_green}{HTML}{b8d4ac}
\begin{frontmatter}

\title{Computational Intelligence based Land-use Allocation Approaches for Mixed Use Areas}

\author[1]{Sabab Aosaf}
\author[2]{Muhammad Ali Nayeem}
\author[3]{Afsana Haque}
\author[1]{M Sohel Rahman\corref{cor1}}
\cortext[cor1]{Corresponding author}
\ead{msrahman@cse.buet.ac.bd}
\address[1]{Department of Computer Science and Engineering, BUET, Dhaka-1000, Bangladesh}
\address[2]{Department of Computer Engineering, College of Computer, Qassim University, Buraydah-51452, Saudi Arabia}
\address[3]{Department of Urban and Regional Planning, BUET, Dhaka-1000, Bangladesh}
\begin{abstract}
Urban land-use allocation represents a complex multi-objective optimization problem critical for sustainable urban development policy. This paper presents novel computational intelligence approaches for optimizing land-use allocation in mixed-use areas, addressing inherent trade-offs between land-use compatibility and economic objectives. We develop multiple optimization algorithms, including custom variants integrating differential evolution with multi-objective genetic algorithms. Key contributions include: (1) CR+DES algorithm leveraging scaled difference vectors for enhanced exploration, (2) systematic constraint relaxation strategy improving solution quality while maintaining feasibility, and (3) statistical validation using Kruskal-Wallis tests with compact letter displays. Applied to a real-world case study with 1,290 plots, CR+DES achieves 3.16\% improvement in land-use compatibility compared to state-of-the-art methods, while MSBX+MO excels in price optimization with 3.3\% improvement. Statistical analysis confirms algorithms incorporating difference vectors significantly outperform traditional approaches across multiple metrics. The constraint relaxation technique enables broader solution space exploration while maintaining practical constraints. These findings provide urban planners and policymakers with evidence-based computational tools for balancing competing objectives in land-use allocation, supporting more effective urban development policies in rapidly urbanizing regions.
\end{abstract}

\begin{keyword}
Decision support system \sep Genetic algorithm \sep Land use planning \sep Optimization \sep Computing
\end{keyword}
\end{frontmatter}


\section{Introduction}
Urban land-use allocation represents a critical operations research problem that fundamentally shapes sustainable urban development worldwide. As global urbanization accelerates, decision-makers face increasingly complex resource allocation challenges involving multiple conflicting objectives, diverse stakeholder preferences, and stringent spatial and economic constraints. This multi-objective optimization problem requires sophisticated analytical approaches to balance competing demands from property developers, urban planners, local communities, and environmental regulators while optimizing land-use compatibility and economic efficiency. The complexity is further compounded by the need to accommodate rapid population growth, ensure disaster resilience, and maintain economic viability within limited urban spaces. These challenges are particularly acute in rapidly developing megacities like Dhaka, Bangladesh, where suboptimal land-use decisions can have far-reaching consequences for millions of residents and economic development trajectories.

Recent fire-related disasters in Dhaka \cite{Rahman2025,BBC2024,DhakaTribune2024} underscore the critical importance of evidence-based land-use allocation decisions for urban safety and resilience. These incidents highlight how inadequate spatial planning can amplify disaster risks, emphasizing the need for robust decision support systems that can systematically evaluate trade-offs between competing objectives while ensuring safety constraints. Advanced operations research methodologies, particularly multi-objective optimization algorithms, have emerged as powerful tools for addressing these complex planning challenges. However, existing approaches often struggle with the combinatorial complexity of real-world land-use allocation problems, particularly when dealing with large-scale urban areas involving thousands of plots and multiple stakeholder objectives. This research addresses these methodological gaps by developing novel computational intelligence approaches that integrate differential evolution concepts with multi-objective optimization frameworks, while introducing systematic constraint relaxation strategies to enhance solution quality without compromising feasibility.

This research makes several significant methodological and practical contributions to the operations research literature on urban land-use optimization:

\begin{itemize}
\item \textbf{Novel algorithmic framework}: We develop a custom multi-objective optimization algorithm (CR+DES) that integrates differential evolution concepts with genetic algorithms, leveraging scaled difference vectors to enhance search space exploration and convergence properties.
\item \textbf{Systematic constraint relaxation methodology}: We introduce and evaluate an approach to constraint relaxation during optimization that maintains solution feasibility while expanding the feasible search space, providing new insights into the trade-offs between constraint adherence and solution quality.
\item \textbf{Real-world validation}: We demonstrate the practical effectiveness of our approaches on a case study involving 1,290 plots in Dhaka's Dhanmondi area, providing empirical evidence of performance improvements and practical applicability.
\item \textbf{Decision support insights}: We provide evidence-based algorithm selection guidance for urban planning practitioners and policymakers based on optimization objectives, supporting more informed land use policy decisions.
\end{itemize}

The rest of the paper is organized as follows. Section~\ref{sec:lit} provides a detailed review of related literature. Section~\ref{sec:pbr} outlines the problem formulation and its constraints. Section~\ref{sec:m} describes our methodology, including solution representation, initialization, optimization framework, and algorithms. Section~\ref{sec:rec} presents the results obtained from all algorithms, and Section~\ref{sec:d} discusses their implications in the broader context of the research. Finally, Section~\ref{sec:n} concludes the paper.

\section{Literature Review} \label{sec:lit}
In this section, we review the related literature in two different segments as follows. We first consider the literature on land-use allocation, with an initial focus on the perspective of urban planners. This stage encompasses a thorough discourse on previous research efforts dedicated to deciphering the intricacies inherent in the land-use allocation. Shifting gears, we then venture into the domain of optimization algorithms, an essential cornerstone for navigating the complex challenges entwined with land-use allocation and urban planning holistically.

\subsection{Land-use Allocation}
As early as 1826, efforts to address land-use allocation challenges, specifically pertaining to agricultural land, were documented \cite{Dauphine2017}, which laid the foundation for contemporary land-use allocation theories. For urban contexts, initial urban land-use allocation models included the Concentric Zone Model \cite{Reiffenstein2017}, the Sector Model \cite{Schwirian2007}, and the Multiple Nuclei Model \cite{Schwirian2007,Harris1945}. These models focused on the interplay between the Central Business District and transportation routes. Later, the introduction of the Bid Rent function \cite{Kirwan1966} provided insights into how plot prices vary with distance from the central business district. Lowry’s Garin–Lowry model first mapped service and residential patterns in Pittsburgh \cite{Fotheringham1985}. A multi-objective LP then balanced conflicting land-use goals across 147 Du Page County regions, halving required acreage \cite{Bammi1976,Bammi1979}. Integrating GIS with multi-criteria evaluation informed optimal UK radioactive waste siting \cite{Carver1991}. Subsequent work addressed competing spatial objectives and trade-offs in land-use allocation \cite{Eastman1998,Beinat1998,Masoomi2013}.

A notable milestone in land-use allocation techniques was reached through the optimization of the Multi-facility location model \cite{Church1999} utilizing the p-median model. This innovative approach, applied to greenfield areas, addressed activity allocation in a novel manner. However, certain studies highlighted limitations, such as the absence of existing land-use patterns at initialization \cite{Ligmann2005}, prompting the proposition of a Zero-one programming model \cite{Williams2002} for land acquisition problems. The latter model incorporated selection criteria encompassing spatial contiguity, total cost, and total area. GIS continued to play a significant role, exemplified in studies seeking optimal sites by maximizing both total costs and total benefits \cite{Li2009}. All the aforementioned methods adhered to linear constraints and involved the allocation of a single land-use to each parcel.

In recent times, heuristic approaches have gained traction for addressing land-use allocation issues with multiple objectives. Simulated Annealing \cite{Semboloni2004} and Particle Swarm Optimization (PSO) \cite{Ma2011} were employed to optimize facility placement within residential and commercial zones, considering factors, such as maximum suitability and the cost of altering land shape. Furthermore, Goal Programming was applied for multi-objective optimization in the context of Tongzhou Newtown in Beijing, China \cite{Cao2012}. Recent developments have seen the realization that linear combinations of objectives may not yield non-convex optimal solutions \cite{Masoomi2013}, leading to the utilization of Pareto Front-based methods. These methodologies demonstrated the capacity to achieve optimal urban land-use arrangements, exemplified in the district of Tehran. Gray multi-objective optimization (GMOP) and patch-generating land-use simulation (PLUS) models were coupled to optimize land-use allocation \cite{Meng2023}. Also, GA and PLUS models were coupled to solve similar problems \cite{Li2022}. Multi-objective optimization algorithm, Non-dominated Sorting Genetic Algorithm II (NSGA-II) was exploited \cite{Mehari2023} for optimizing urban land-use allocation. Optimization was coupled with territorial LCA for agricultural land-use planning \cite{Li2022}. Recently, a spatially rationalized framework was introduced by coupling hedonics pricing model and genetic algorithm \cite{MEHARI2025107694}, and a spatially explicit multi-objective optimization tool for green infrastructure planning was developed based on InVEST and NSGA-II \cite{DONG2025107465}.

\subsection{Optimization}
The landscape of multi-objective optimization techniques for urban land-use allocation has evolved through intricate methodologies. A significant breakthrough was the introduction of a vector-evaluated Genetic Algorithm (GA) that inherently tackled multiple objectives \cite{Schaffer1985}. The Weight-based Genetic Algorithm, coupled with Multi-Objective Evolutionary Algorithms (MOEAs) \cite{Mauledoux2015}, underscores the efficient handling of competing objectives in urban planning scenarios. Additionally, the Niched Pareto Genetic Algorithm \cite{Horn1994,Horn1993}, Strength Pareto Evolutionary Algorithm 2 (SPEA2) \cite{Sheng2012}, NSGA, and Fast NSGA-II \cite{Deb2000} contributed to the complex puzzle of optimizing multiple objectives in land-use allocation. Critical to this discourse is the empirical validation of these methodologies, a phase realized through meticulous evaluation in diverse scenarios \cite{Fonseca1993,Deb2002,Mauledoux2015}. 

Intriguingly, the evolution of multi-objective optimization extends beyond GAs. The fusion of diverse GA methods is evident \cite{Konak2006}, while GAs find application in catering to both planners and real estate developers in urban land-use allocation \cite{Haque2014}. The integration of NSGA-II in generating alternative land-use allocations through a multi-objective optimization approach was also demonstrated \cite{Sharmin2019}, further amplifying the significance of these techniques. To the best of our knowledge, Differential Evolution (DE) \cite{Storn1997}, another variant of genetic algorithms, has not been used for land-use allocation. Optimizing its parameters, like the scale factor ($F$), is crucial as striking the right balance \cite{Ali2007} is pivotal to avoid efficiency pitfalls. However, excessively increasing the scale factor ($F$) can lead to a uniform probability distribution for generated points \cite{Wang2010}, hampering the effectiveness of mutation operators. The DE refinement journey extends further through analyses of crossover operations \cite{Zaharie2009} and innovative control parameter strategies, as seen in the self-adapting jDE \cite{Pan2011}. Advancements in self-adaptation strategies \cite{Qin2009} reveal the depth of research-driven optimization exploration. The continuum of progress culminates in the utilization of adaptive differential evolution with ensemble operators for continuous optimization problems \cite{Yi2022}.

\section{Problem Formulation} \label{sec:pbr}
We inspect an already developed area comprising $N$ plots (a plot is a piece of well marked land where a building is constructed) of different shapes and sizes. Buildings of different stories are situated in these plots. Each of the stories of these buildings may have one or more land-uses, like residential, commercial, office, and so on. There may be a total of $K$ types of such land-uses. The problem is to determine the allocation of $p$ ($p = 1, 2, ... K$) land-uses in the buildings of $N$ plots in such a way that best achieves the objectives. The decision variable $x_{i,m}$ is defined as the proportion of total floor space of plot $i$ taken by land-use $m$. The problem formulation is borrowed from \cite{Sharmin2019}.

To solve the land allocation problem, two objectives are considered: maximizing compatibility and maximizing price. Compatibility is defined by Equation \ref{eq:compatibility}, and the price is defined by Equation \ref{eq:price}. 

\begin{equation}
\text{Compatibility} = \sum_{i \in I} \sum_{j \in J(i)} \sum_{l=1}^{K} \sum_{m=1}^{K} C_{l,m} \cdot x_{i,l} \cdot x_{j,m} \cdot F_i \cdot F_j
\label{eq:compatibility}
\end{equation}

Here, $K$ = number of types of land-uses; $I$ = set of plots. $J(i)$ = neighborhood of plot $i$. $C_{l,m}$ = compatibility index of land-use $l$ and $m$. $x_{i,m}$ = proportion of use m in plot $i$. $x_{j,l}$ = proportion of use $l$ in plot $j$. Here, the proportion of a use is the ratio of the floors needed for a specific use and the total number of floors. $F_i$ = total floor space of plot $i$ and, $F_j$ = total floor space of plot $j$. The presence of variable $x_{i,l}$ in Equation \ref{eq:compatibility} gives rise to non-linearity in the first objective function. $F_i$ and $F_j$ capture the degree of compatibility in the formulation with the assumption that the compatibility is proportional to the area under a certain use. 

\begin{equation}
\text{Price} = \sum_{i=1}^{N} \sum_{m=1}^{K} P_{i,m} \cdot x_{i,m}
\label{eq:price}
\end{equation}

Here, $K$ = number of types of land-uses. $i$ = plot number and $m$ is the use. Also, $P_{i,m}$ = price of plot $i$ for use $m$. $x_{i,m}$ = proportion of use $m$ in plot $i$. 

\subsection{Constraints} \label{sec:cn}
A valid solution of the problem is subject to the following four constraints.

\textbf{Constraint 1.}
\begin{equation}
\sum_{m=1}^{K} x_{i,m} = 1 \quad \forall i \in I
\label{eq:constraint1}
\end{equation}

Here, $K$ = number of types of land-uses. $x_{i,m}$ is the proportion of total floor space of plot $i$ taken by land-use $m$. The summation has to be 1.

\textbf{Constraint 2.}
\[ 0 \leq X_{im} \leq 1 \]

\textbf{Constraint 3.}
\begin{equation}
(1-\gamma) \cdot \sum_{i=1}^{N} A_{i,m}^{\text{actual}} \leq \sum_{i=1}^{N} x_{i,m} \cdot F_i \leq (1+\gamma) \cdot \sum_{i=1}^{N} A_{i,m}^{\text{actual}}
\label{eq:constraint3}
\end{equation}

Here, $N$ = Number of Plots. $F_i$ is the total floor area of plot $i$. $A_{i,m}$ is the total floor area of land-use $m$ for plot $i$. The summation gives us the total floor area taken by land-use $m$ in the study area. This is calculated for the solution suggested by the optimization algorithm. $A_{i,m}^{\text{actual}}$ is the floor area of land-use m for the actual land-use. And $\gamma$ is the percentage of change allowed (referred to as the Area Constraint). So for a particular percentage 10\% we allow 10\% decrease to 10\% increase for the floor area of a particular land-use.

\textbf{Constraint 4.}
\begin{equation}
P_{\text{min}} \leq \sum_{i=1}^{N} \sum_{m=1}^{K} P_{i,m} \cdot x_{i,m} \leq P_{\text{max}}
\label{eq:constraint4}
\end{equation}

Here, $K$ = number of types of land-uses. $N$ = Number of Plots. $P_{i,m}$ is the price of the portion of floor area of the plot $i$ allocated by land-use m. $\sum_{i=1}^{N} \sum_{m=1}^{K} P_{i,m} \cdot x_{i,m}$ gives us the total price of the study area. We give lower and upper price limits and keep the price of the area in between. $P_{\text{min}}$ is the lower limit of price and $P_{\text{max}}$ is the upper limit of price.

To guide the optimization process, we may introduce another constraint (Constraint 5).

\textbf{Constraint 5.}
\begin{equation}
\sum_{i=1}^{N} y_i \leq \mu \cdot N
\label{eq:constraint5}
\end{equation}

Here, $N$ = Number of Plots. $y_i$ is a boolean variable. Its value is 1 if land-use of plot $i$ is changed and 0 otherwise. $N$ is the total number of plots. $\mu$ is the percentage of plots that are allowed to get their land-use changed (Plot Constraint). The end results may not satisfy this constraint. Certain designated plots, such as those allocated for educational Institutes and hospitals, are not altered due to their sensitive nature. Such plots are excluded from consideration in the optimization algorithms.

\section{Methods} \label{sec:m}
This section presents our algorithmic approaches, delineating our strategies across four distinct dimensions, namely, Single Objective Approach (SOA), Non-dominated Sorting Genetic Algorithm II (NSGA-II), Two Archive Approach (C-TAEA, which is a recent MOEA), and custom algorithms (MSBX+MO and CR+DES).

\subsection{General Structure}
We use several optimization algorithms in this paper. However, all of them have the following three steps.

\textbf{Step 1:} We first generate a representation for actual land-use allocation of our study area. (details in Section \ref{sec:rep})

\textbf{Step 2:} Then, from the actual land-use representation, we generate more solutions by applying a constrained random change to some of the land-uses (i.e., not changing more than the given constraints). We generate 100 land-use allocation representations. (details in Section \ref{sec:i})

\textbf{Step 3:} We use operators (details in Section \ref{sec:op}) to generate new solutions or move solutions (towards optimal location) in the solution space for a certain number of iterations and choose the best solutions.

Step 1 and 2 are the same for all optimization algorithms we have in the study, with possible differences in Step 3.

\subsubsection{Solution Representation} \label{sec:rep}
\begin{figure}[!htbp]
\centering
\includegraphics[width=1\textwidth]{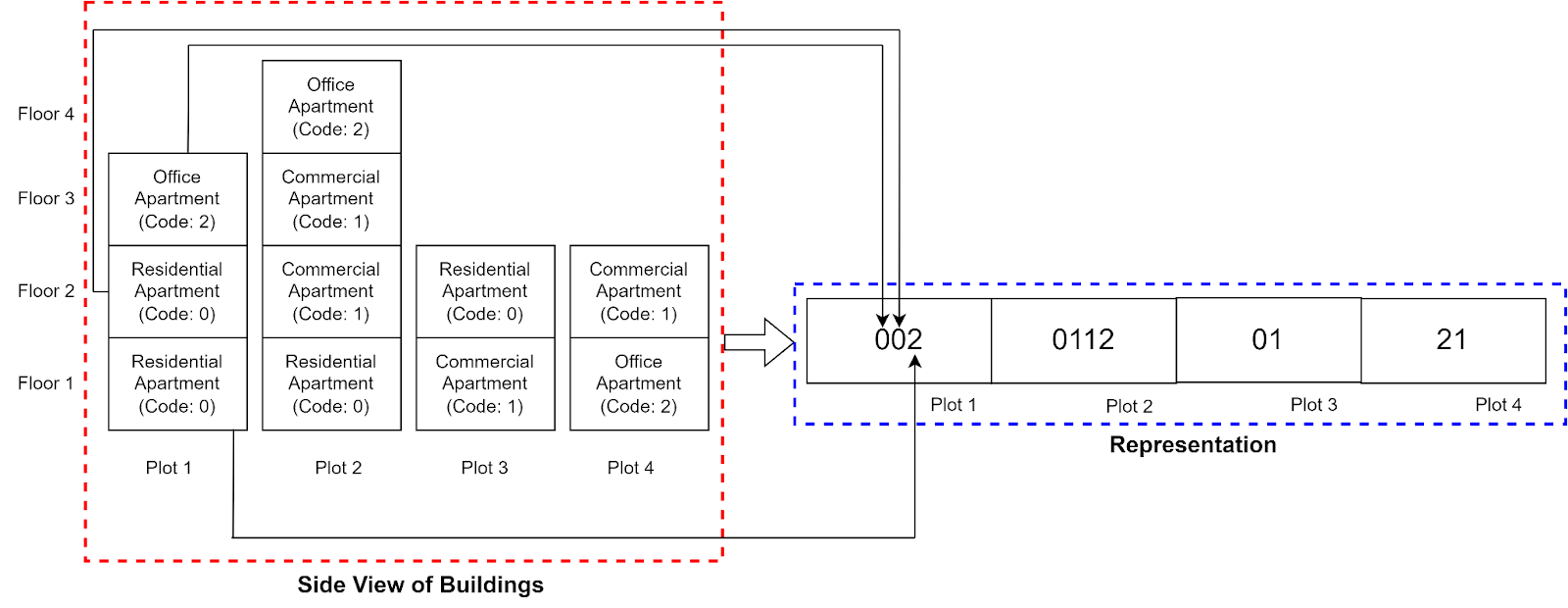}
\caption{An instance of land-use allocation and corresponding representation is shown in this figure. In the red dashed rectangle, we represent the buildings on a plot, and inside the blue dashed rectangle, we have the corresponding representation. For each plot, there can be a building with multiple floors with different land-uses. The codes representing the land-use of a plot are kept in the representation array.}
\label{fig:representation}
\end{figure}

We demonstrate our representation scheme with the help of a simplified instance in Figure \ref{fig:representation}, where we have shown an instance with 4 plots for allowing land-use change in an area (our study area has 1290 such plots). In this instance, the building situated on plot 1 has 3 floors, the building situated on plot 2 has 4 floors, and so on. The land-use of each floor of a building (in the left), along with the code (in the right) for the particular land-use, is also shown, where we assigned the following coding scheme: residential flat = 0, commercial = 1, and office = 2. Inside the dashed blue rectangle, we show the representation of the solution, which is an array of strings composed of characters 0, 1, and 2, following the land-use coding scheme.

\subsubsection{Generating Initial Population} \label{sec:i}
We generate 100 initial solutions. Typically, the entire population is initialized randomly to ensure maximum diversity among the solutions. However, acknowledging the complexity of producing valid solutions abiding by all constraints (as defined in Section \ref{sec:cn}), a complete randomization is avoided. More specifically, in our approach, each of the 100 solutions, or representations (equivalent to 100 land-use maps of the study area), is crafted through altering the actual land-use configuration of 25\% of the plots, thereby ensuring a diversified set of starting points and encapsulating the variability inherent in land-use allocation scenarios. We adopted this idea from \cite{Sharmin2019}.

\subsubsection{Optimization Operators} \label{sec:op}
The initial phase of a metaheuristic-based optimization process involves a selection procedure. Then the chosen individuals undergo crossover and/or mutation operations with the hope to get better individuals.

\textbf{Selection}\\
The process of selecting individuals for crossover is executed through a tournament selection mechanism as follows. From the population, two individuals are chosen at random, and the individual with the higher fitness level is chosen to enter the mating pool. This selection process is iterated until the mating pool reaches its designated size of half the population. This approach ensures that individuals with better fitness have a higher chance of being included in the mating pool, promoting the propagation of more favorable traits across generations.

\textbf{Mutation}\\
In some cases, we perform mutation on a candidate solution before crossover. Also, instead of doing crossover every time, with some probability, we choose only mutation sometimes. Depending on the algorithm of our choice, the probability (i.e., mutation rate) of applying mutation may differ.

\textbf{Crossover}\\
We employ crossover operators in our optimization algorithms to enable search space exploitation. Our crossover operates on two parent solutions and generates two offspring solutions. We adapt two crossover operators, SBX (Simulated binary Crossover) \cite{deb1995simulated} and Uniform Crossover, to our requirements discussed as follows.

\begin{figure}[!htbp]
\centering
\includegraphics[width=1\textwidth]{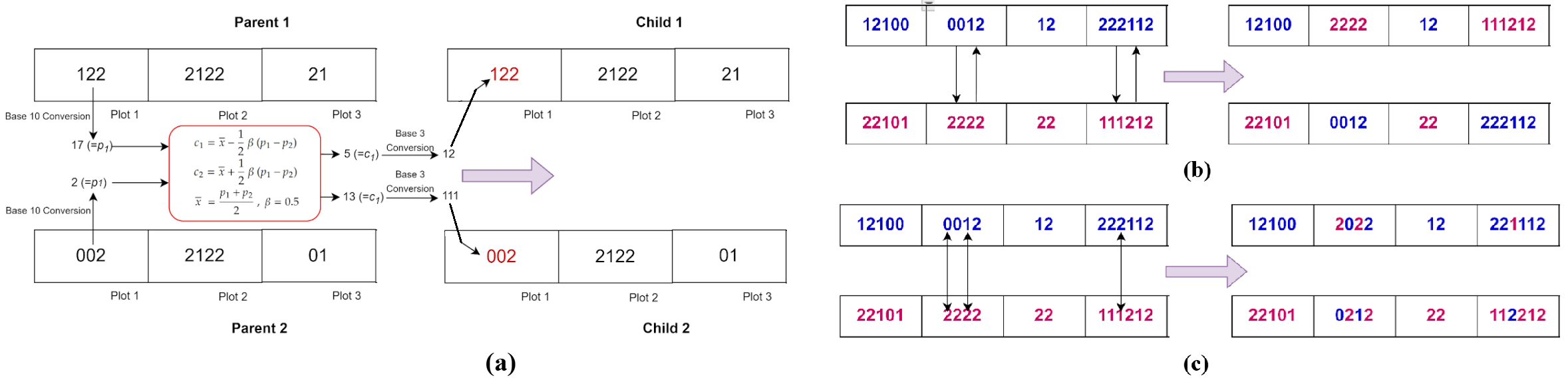}
\caption{(a) We show two parents and two generated children using SBX crossover. Two corresponding plots of the two parents are taken. And using the crossover operation, two new values are generated, which are then put into the children. The crossover is done for multiple corresponding plots. Here we show a crossover for one plot. (b) Uniform crossover swapped the land-use of plots to get new solutions. (c) Uniform crossover swapped the Land-use for the same floors of two solutions.}
\label{fig:crossover}
\end{figure}

\textit{SBX (Simulated Binary Crossover):} We demonstrate our SBX Crossover method through a simplified instance in Figure \ref{fig:crossover}(a). First, two random individuals are selected as parents (Parent 1 and Parent 2 in the Figure). For selected cells (plots) the ternary values representing the land-uses (of building on that plot) are converted to a base 10 integers before SBX Crossover (in Figure \ref{fig:crossover}(a), the '122' value of the first cell is converted to 17). The SBX operator generates two new land-uses from a plot (Plot 1) from the parents, which are then put into two children. After the SBX Crossover, the values of the cells are converted back to ternary values. In Figure \ref{fig:crossover}(a), we demonstrate crossover of only one plot (Plot 1) for simplicity, but there can be multiple plots.

\textit{Uniform Crossover:} For some of our algorithms, we use Uniform Crossover. With a pre-specified probability, land-uses of some of the plots are swapped with the land-uses of the same plots of another solution. This way we get two children from two parents as illustrated in Figure \ref{fig:crossover}(a) and \ref{fig:crossover}(b). In some runs we also apply the crossover floor wise by swapping floor uses of two solutions (Figure \ref{fig:crossover}(c)).\\

\subsection{Optimization Algorithms}
In this section, we will present a detailed explanation of our optimization algorithms. The algorithms are SOA (Single Objective Approach), Mutation+SBX NSGA-II, Two Archive Approach, and two novel custom approaches.
\subsubsection{SOA (Single Objective Approach)}
\begin{figure}[!htbp]
\centering
 \includegraphics[width=1\textwidth]{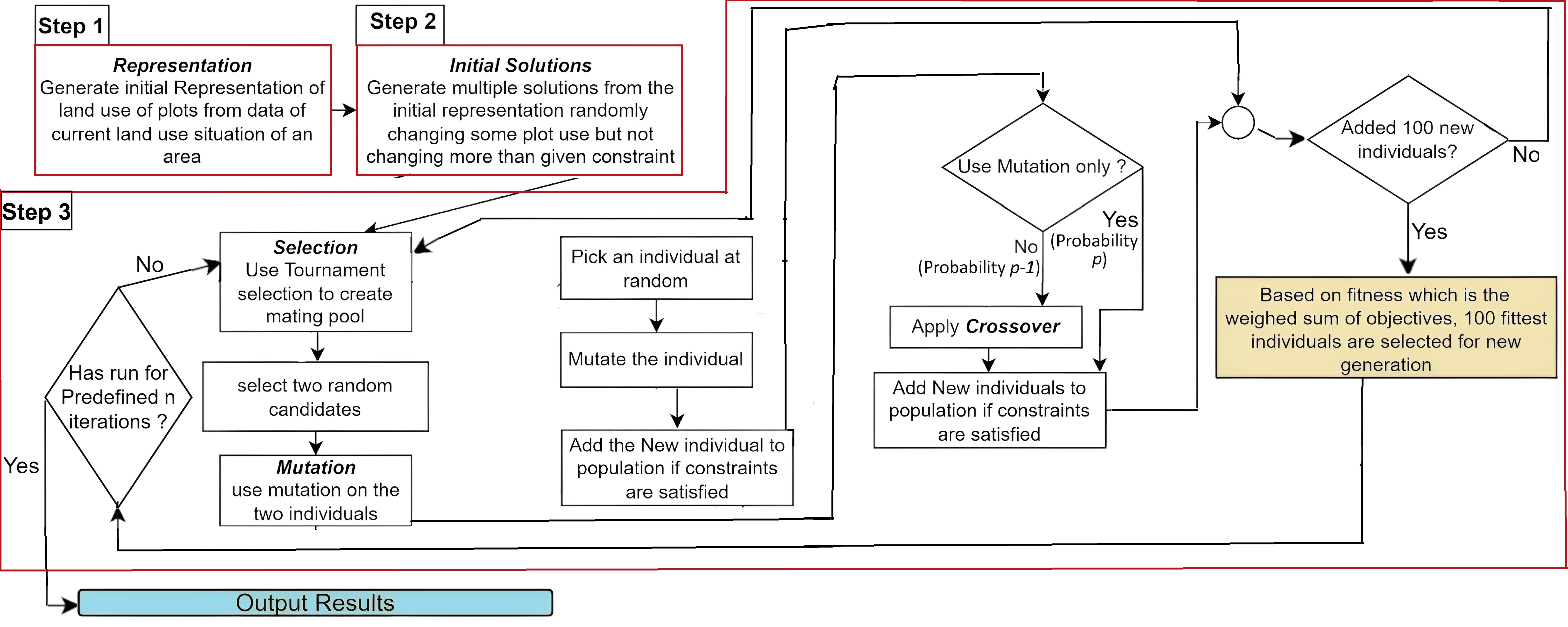}
\caption{Flowchart of SOA.
}
\label{fig:soa_flowchart}
\end{figure}

Recall that we have two objectives: price and compatibility. In the SOA, we merge these two objectives into one using a linear equation. The coefficients or weights for price and compatibility are determined by running the SOA multiple times with different coefficient values. In Figure \ref{fig:soa_flowchart}, we show the steps of SOA. There are three steps.

\textbf{Step 1:} Representation is generated from actual data. (descriptions are in Section \ref{sec:rep})

\textbf{Step 2:} Initial population is generated. (descriptions are in Section \ref{sec:i})

\textbf{Step 3:} Next, from this population, we choose the mating pool through tournament selection, where the fitness is determined by a weighted sum of objective functions. Following this, we introduce mutations or crossovers, or a combination of both, based on probability, and incorporate the resulting offspring into the population. Subsequently, we select the fittest individuals to form the next generation. Once we have the new generation, we repeat the process starting from Step 3.

In Table \ref{tab:configuration} (Column 1), the fitness calculation equation for SOA approach and configurations are specified.

\begin{table}[!htbp]
\footnotesize
\centering
\caption{Configuration of SOA, Mutation+ SBX NSGA-II and Two Archive Optimization Algorithms}
\begin{tabular}{p{3cm}p{2cm}p{2cm}p{3cm}p{4cm}}
\toprule
Algorithm & \begin{tabular}{@{}l@{}}Mutation\\before\\crossover\end{tabular} & Crossover & \begin{tabular}{@{}l@{}}Mutation\\or other\\approach\\(In place\\of crossover)\end{tabular} & Fitness \\
\midrule
SOA & Random Mutation & SBX & Random Mutation & $a \times \text{price} + b \times \text{compatibility}$ \\ \cdashlinelr{1-5}
\begin{tabular}{@{}l@{}}Mutation+ SBX\\NSGA-II\end{tabular} & Random Mutation & SBX & Polynomial mutation & \begin{tabular}{@{}l@{}}Pareto rank and\\crowding distance\end{tabular} \\ \cdashlinelr{1-5}
TwoArc & Random Mutation & Uniform crossover & N/A & \begin{tabular}{@{}l@{}}Convergence\\and Diversity\end{tabular} \\
\bottomrule
\end{tabular}
\label{tab:configuration}
\end{table}

\subsubsection{Mutation+SBX NSGA-II}
We use an NSGA-II based algorithm inspired by \cite{Sharmin2019}, which used mutation before Crossover (and after). We name our NSGA-II based method Mutation+SBX NSGA-II. Mutation+SBX NSGA-II, unlike NSGA-II \cite{Sharmin2019}, incorporates a crossover step after mutation. The steps of Mutation+SBX NSGA-II are as follows.

\textbf{Step 1:} Representation is generated from actual data. (as detailed in Section \ref{sec:rep})

\textbf{Step 2:} Initial population is generated. (as detailed in Section \ref{sec:i})

\textbf{Step 3:} From this population, we employ tournament selection with Pareto rank as the fitness criterion to choose individuals for a mating pool. We then apply mutation, crossover, or a combination of both with certain probabilities and introduce the resulting offspring into the population. Following this, we identify the most fit individuals for the next generation using Pareto Fronts and crowding distance. Consequently, we obtain a new generation, and the process returns to the beginning of Step 3.

In Table \ref{tab:configuration}, we show the configuration of the Mutation+SBX NSGA-II algorithm. The fitness is the Pareto rank and the crowding distance (for breaking ties, where Pareto ranks are the same). We also used several other variants of NSGA-II that are mentioned in Supplementary Section S1.\\

\subsubsection{The Two Archive Approach}
We implement a two archive optimization algorithm named C-TAEA \cite{Shan2021} (two-archive evolutionary algorithm for constrained multi-objective optimization). The C-TAEA approach guides the optimization process by using a convergence archive (CA) and a diversity archive (DA). This approach of dual archiving balances convergence and diversity for C-TAEA. The specifications of Two Archive approach are given in Table \ref{tab:configuration}.\\

\subsubsection{Customized Novel Algorithms}
We develop two customized algorithms, MSBX+MO and CR+DES (which hybridizes two classical methods, DE with NSGA-II). We describe them as follows.

\textit{MSBX+MO:} Inspired by DE, we develop the strategy, MSBX-MO, where for each solution x, the scaled value of another random solution is directly added to x to get the mutant. The mutation is crossed over with x to get children. In Table \ref{tab:custom_algorithms}, the mutation strategy of MSBX+MO is shown. It uses the same crossover operator and fitness measurement, albeit with a different mutation strategy. Along with this approach, we also implement traditional DE and some customized DE methods (The details are in Supplementary Sections S1).

\textit{CR+DES:} Inspired by NSGA-II and DE, we develop this multi-objective algorithm that allows scaled difference vectors to be directly inserted into the candidate solution list. In each iteration, we generate children by performing crossover and also with a predefined probability we generate some children by taking the scaled difference of two solutions. Other components of the algorithm are same as NSGA-II. In Table \ref{tab:custom_algorithms}, the configuration of CR+DES is reported.

\begin{table}[h]
\footnotesize
\centering
\caption{Configuration of MSBX+MO and CR+DES}
\begin{tabular}{lp{5cm}lp{3.5cm}}
\toprule
Algorithm & Customization & Crossover & Fitness \\
\midrule
MSBX+MO & Custom Mutation Strategy Employed & SBX & Pareto rank \\ \cdashlinelr{1-4}
CR+DES & Difference vector used in DE is used as the candidate solution. This was done in place of crossover with a probability & Uniform crossover & Pareto rank and crowding distance \\
\bottomrule
\end{tabular}
\label{tab:custom_algorithms}
\end{table}
\subsection{Evaluation Metrics}
Since our problem is multi-objective, we evaluate each algorithm by analyzing both the Pareto front it generates and the best-achieved values for the two objectives. In addition to visual inspection, we employ several quantitative performance indicators specifically designed to assess the quality of the Pareto fronts \cite{Blank2020}. The indicators used in this study include the Hypervolume Indicator (HV) \cite{Guerreiro2021}, Generational Distance (GD) \cite{Riquelme2015}, Generational Distance Plus (GD+) \cite{Riquelme2015, Ishibuchi2019}, Inverted Generational Distance (IGD) \cite{Ishibuchi2019}, and Inverted Generational Distance Plus (IGD+) \cite{Ishibuchi2019}.

The GD metric measures the distance from solution to the Pareto-front. Let us assume the points found by our algorithm are the objective vector set A=$\{a_1, a_2, ..., a_n\}$ and the reference points set (Pareto-front) is Z=$\{z_1, z_2, ..., z_m\}$. The equation \cite{Blank2020} to compute GD is given below.

\begin{equation}
GD(A,Z) = \frac{1}{|A|} \left(\sum_{i=1}^{|A|} d_i^p\right)^{1/p}
\label{eq:gd}
\end{equation}

Here, $d_i$ represents the Euclidean distance (p = 2) from $a_i$ to its nearest reference point in Z. Basically, this results in the average distance from any point A to the closest point in the Pareto-front.

The GD+ metric is based on the GD metric, which is also used to evaluate how close a set of solutions is to the true Pareto front. The equation \cite{Blank2020} to compute GD+ is given below.

\begin{equation}
GD^+(A,Z) = \frac{1}{|A|} \left(\sum_{i=1}^{|A|} (d_i^+)^p\right)^{1/p}
\label{eq:gdplus}
\end{equation}

Here, $d_i^+ = \max\{a_i - z_i, 0\}$ represents the modified distance from $a_i$ to its nearest reference point in Z with the corresponding value $z_i$.

The IGD performance indicator inverts the generational distance and measures the distance from any point in Z to the closest point in A. The IGD+ metric is based on the IGD metric \cite{Blank2020}.

\begin{equation}
IGD(Z,A) = \frac{1}{|Z|} \left(\sum_{i=1}^{|Z|} \hat{d}_i^p\right)^{1/p}
\label{eq:igd}
\end{equation}

\begin{equation}
IGD^+(Z,A) = \frac{1}{|Z|} \left(\sum_{i=1}^{|Z|} (\hat{d}_i^+)^p\right)^{1/p}
\label{eq:igdplus}
\end{equation}

Here, $d_i$ represents the euclidean distance (p = 2) from $z_i$ to its nearest reference point in A. $d_i^+ = \max\{z_i - a_i, 0\}$ represents the modified distance from $z_i$ to the closest solution in A with the corresponding value $a_i$.

The HV indicator calculates the area/volume, which is dominated by the provided set of solutions with respect to a reference point. For maximization problems, a higher value of HV indicator means better performance. 

\subsection{Statistical Tests} \label{sec:stat}
Due to the stochastic nature, we conduct five independent runs for each optimization algorithm. While five runs per algorithm is a small sample, it reflects practical limitations due to each run taking nearly 4-5 hours. We use suitable non-parametric statistical tests at a 0.05\% significance level to compare and assess algorithm performance. We first employ the Kruskal–Wallis test~\cite{kruskal1952use} to determine whether there are statistically significant differences in performance among the algorithms. The Kruskal–Wallis test is appropriate for our setting because:
\begin{itemize}
    \item The data are not assumed to follow a normal distribution.
    \item Each algorithm is evaluated independently.
    \item The number of observations per algorithm (five) is small and unequal variance is possible.
\end{itemize}
This test evaluates the null hypothesis that all algorithms have equal performance distributions. A low \( p \)-value ($\le 0.05$) from this test indicates that at least one algorithm performs significantly differently from the others.

Upon obtaining a statistically significant result from the Kruskal–Wallis test, we conduct post-hoc pairwise comparisons using the Bonferroni-Dunn procedure~\cite{dunn1961multiple}. Bonferroni-Dunn is a rank-based method suitable for identifying which pairs of algorithms differ significantly while controlling for the family-wise error rate in multiple comparisons. This method calculates the differences in average ranks between each pair of algorithms and compares them against a critical value derived from the desired significance level. Finally, we present the results of the pairwise comparisons in a concise and interpretable form, we utilize the compact letter display (CLD) technique. In this approach, each algorithm is assigned a letter (or a set of letters) such that:
\begin{itemize}
    \item Algorithms sharing at least one common letter are not significantly different from each other.
    \item Algorithms that do not share any common letters are statistically different.
\end{itemize}
 The CLD letters are derived from the Bonferroni-Dunn post-hoc test results, where each algorithm's average rank is compared against others. The resulting letter assignments can be displayed graphically~\cite{yang2025metabolite}, allowing for easy identification of statistically equivalent groups and highlighting the best-performing algorithms.

Overall, the combined use of Kruskal–Wallis, Bonferroni-Dunn post-hoc test, and CLD provides a statistically sound and interpretable methodology for comparing the performance of multiple stochastic algorithms under limited evaluation budgets.

\subsection{Code, Environment and Availability}
We use two machines with different configurations for our experiments as follows.

\begin{itemize}
\item Configuration 1: Processor: Ryzen Threadripper 1920x (12 Cores, 3.6 Ghz), Ram: 16 GB, SSD as Storage
\item Configuration 2: Processor: Intel Core i5 (2 cores, 2.9 GHz), Ram: 4GB, SSD as Storage
\end{itemize}

We use MATLAB R2022a, R2023a, and Pycharm (Python 3.7) for implementing the optimization algorithms. With Configuration 1, approximately 4 hours are taken to run each algorithm. On the other hand, with Configuration 2, approximately 12 hours are taken. The resource utilization is: 10\% processor (and 5\% RAM) for Configuration 1 and 15\% processor (and 20\% RAM) for Configuration 2.

\section{Results} \label{sec:rec}
In this section, we present a thorough comparison of various approaches employed in our study, including our customized proposed approaches. Subsequently, we illustrate the outcomes achieved through the manipulation of constraint values. This dual exploration provides valuable insights into the relative performance of different methodologies and the impact of constraint adjustments on the optimization results. For generating pareto front, each method is run five times due to the stochastic nature of the algorithms and combined pareto front from the five runs is analyzed. As performance indicators, the mean of five runs is reported (Table \ref{tab:performance_metrics}).

\begin{table}[h]
\centering
\footnotesize
\caption{Parameters for metaheuristic optimization algorithms. Parameter values in colored regions are changed in the relaxed constrained versions of our algorithms.}
\begin{tabular}{p{1.8cm}p{1.8cm}p{1.8cm}p{1.8cm}p{1.8cm}p{1.8cm}p{1.8cm}}
\toprule
\begin{tabular}{@{}l@{}}Initial\\population\\change\end{tabular} & Population & \begin{tabular}{@{}l@{}}Area\\constraint\\$\gamma$\end{tabular} & \begin{tabular}{@{}l@{}}Plot\\constraint\\$\mu$\end{tabular} & \begin{tabular}{@{}l@{}}Gener-\\-ations\end{tabular} & \begin{tabular}{@{}l@{}}Minimum\\Price\\allowed\\(BDT),\\$P_{\text{min}}$\end{tabular} & \begin{tabular}{@{}l@{}}Maximum\\Price\\allowed\\(BDT),\\$P_{\text{max}}$\end{tabular} \\
\midrule
25\% & 100 & \cellcolor{nice_green}30\% & \cellcolor{nice_yellow}20\% & 150 & \begin{tabular}{l}130000\\000000\end{tabular} & \begin{tabular}{l}145000\\000000\end{tabular} \\
\bottomrule
\end{tabular}
\label{tab:parameters}
\end{table}
In Table \ref{tab:parameters}, we present the parameters of our metaheuristic optimization algorithms. For generating the initial population, the actual land-use map (that is in the dataset) is modified by changing the area of each land-use at most 25\% (also discussed in Section \ref{sec:i}). The population size is 100 for all our optimization algorithms. We also see from \cite{Sharmin2019} that, in general, achieving more significant changes in land-use tends to lead to more optimal outcomes. However, it is important to acknowledge that implementing these changes can be financially demanding. To mitigate the need for extensive redevelopment expenses, we opt to make incremental adjustments, specifically applying a modest percentage (30\% as given in Column 3 of Table \ref{tab:parameters}) of certain space use changes. This constraint corresponds to the area constraint $\gamma$ described in Section \ref{sec:cn}. On the other hand, 20\% of plots are allowed to change their actual land-use (Column 3 of Table \ref{tab:parameters}), so that the new solutions generated by the optimization procedure are more likely to satisfy constraints. This corresponds to plot constraint $\mu$, described in Section \ref{sec:cn}.

The upper and lower limits of price are found through a trial-and-error approach based on the existing land prices in the study area in \cite{Sharmin2019}. After numerous iterations, an upper bound is established by increasing the existing land price by 9.7\% and the lower bound by decreasing it by 1.65\%. A minimum reduction of 1.65\% from the baseline land price is set, considering the trade-off between area compatibility and land price. We keep the price in the range 130 billion to 145 billion BDT (price constraint $P_{\text{min}}$ and $P_{\text{max}}$, respectively, as described in Section \ref{sec:cn}). All the optimization algorithms except for the relaxed constrained ones are described in Tables \ref{tab:configuration} and \ref{tab:custom_algorithms}; the latter ones are described in Table \ref{tab:relaxed_constraints}.

\subsection{Comparison of approaches}
As has been mentioned above, we compare our approaches using a 30\% constraint for land-use area change and keep the price between 130 and 145 billion BDT. For all approaches, our population size is 100. The stopping criterion is 150 iterations. Except for SOA, for all approaches, the Pareto Front at the 150th iteration is taken. For SOA, we get the best solution using our fitness function thereof.

\begin{figure}[h]
\centering
\includegraphics[width=0.7\textwidth]{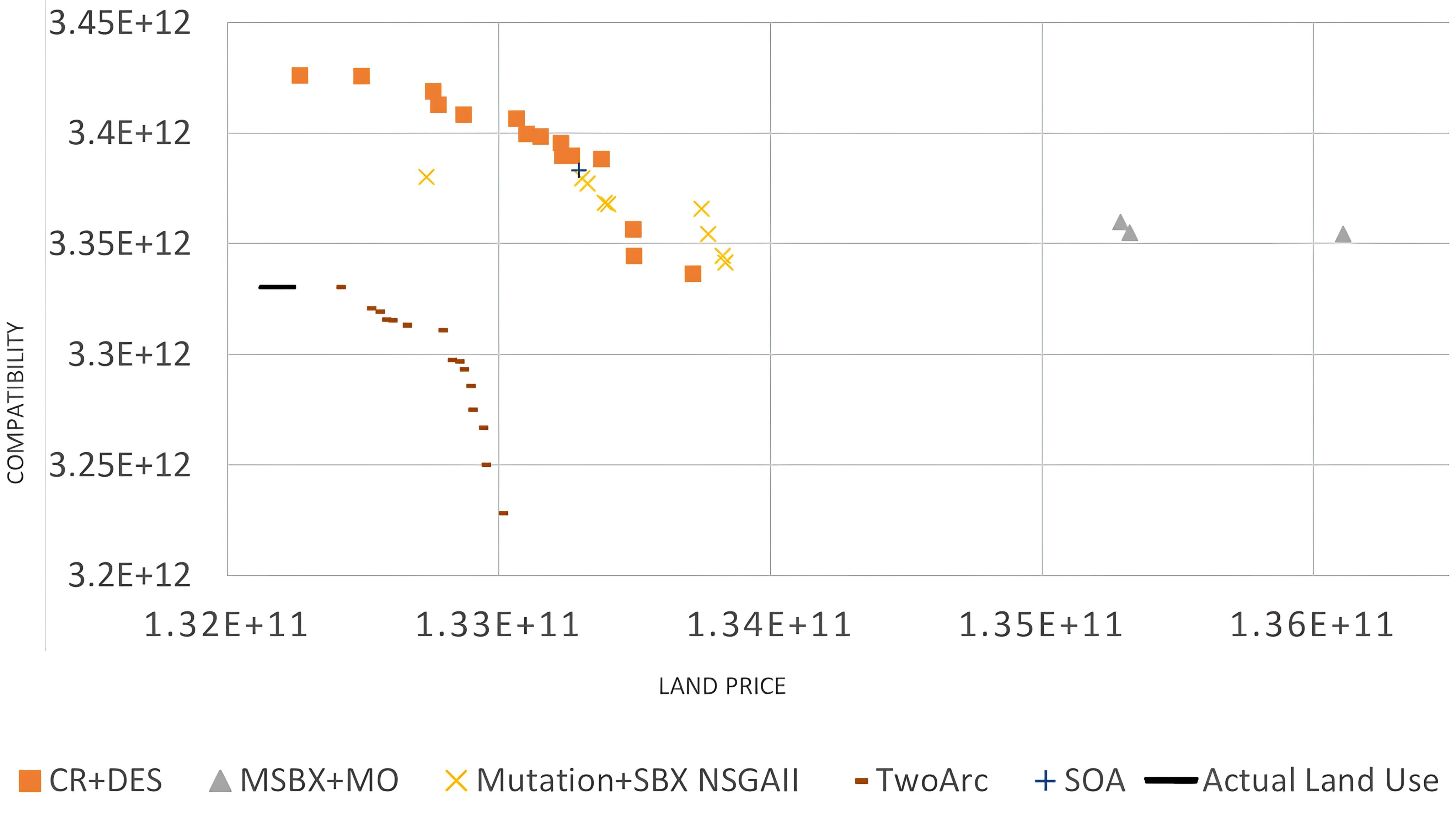}
\caption{Comparison of Pareto Front of different algorithms at the 150th iteration. X-axis has land price, and Y-axis is the compatibility. Some solutions are hidden completely due to overlapping.
}
\label{fig:pareto_comparison}
\end{figure}
In Figure \ref{fig:pareto_comparison}, we compare the Pareto Front of different approaches. We also plot the actual land-use scenario and the Pareto Front we obtain just after initialization. After initialization, we see the solutions of the Pareto Front have better price but worse compatibility than the actual land-use scenario. Out of all solutions of all approaches, the best price is given, but a solution that resides in the Pareto Front of MSBX+MO optimization algorithm. And the solution with the best compatibility is found in the Pareto Front of the CR+DES optimization algorithm. The combined Pareto Front features the solutions of CR+DES, MSBX+MO, and Mutation+SBX NSGA-II. CR+DES has the highest number of solutions in the combined Pareto Front. TwoArc and CR+DES solutions are more diverse as the Pareto Fronts are spread, which may facilitate flexible decision making. But MSBX+MO is less spread out. Also, all TwoArc solutions are dominated by one or more solutions from all other methods, indicating worse performance by TwoArc. As a performance indicator, we calculate the hypervolumes (HV) of the Pareto Fronts (Table \ref{tab:performance_metrics}). Among the methods shown in Figure \ref{fig:pareto_comparison}, MSBX+MO exhibits the highest HV (Table \ref{tab:performance_metrics}).

\subsection{Single objective analysis}
Here, we present the outcomes obtained when transitioning from a multi-objective approach to SOA. Additionally, we provide a comprehensive exploration of how adjustments to the coefficients within the equation, combining multiple objectives, can significantly influence the final outcomes. This deeper examination helps us to better understand the intricate relationship between these coefficients and the results.

\begin{figure}[h]
\centering
\includegraphics[width=1\textwidth]{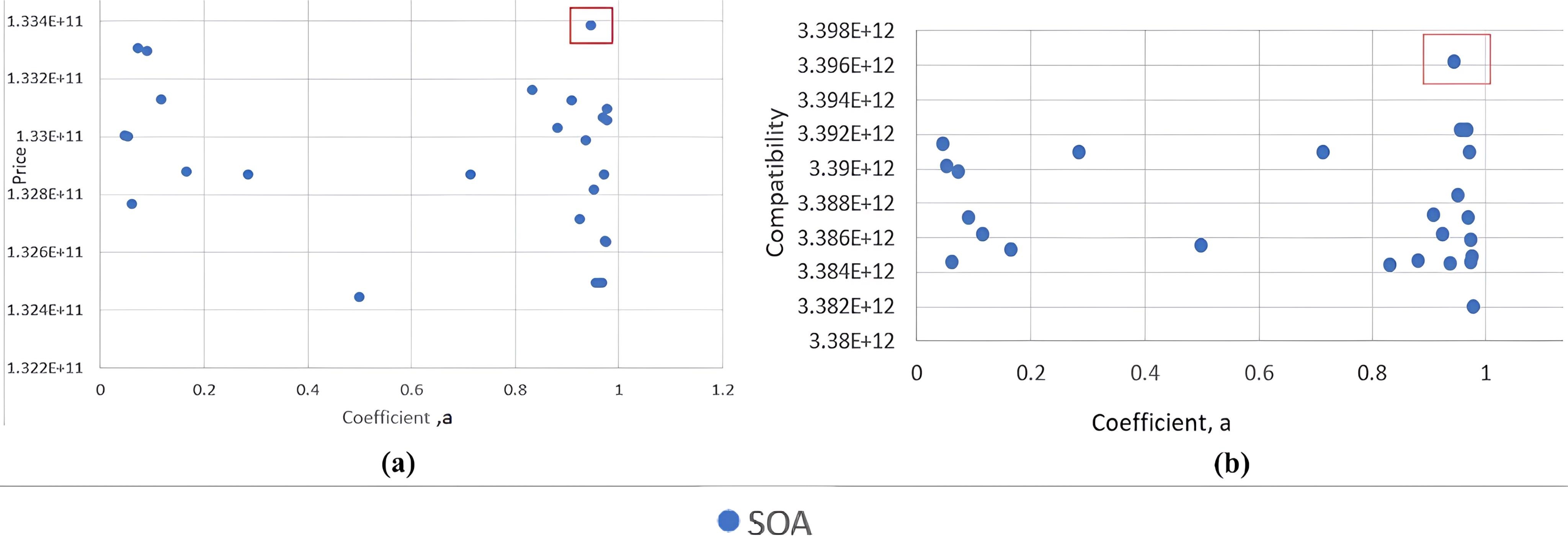}
\caption{(a) Prices of individuals of SOA for different coefficient values at the 150th iteration.
(b) Compatibilities of individuals of SOA for different coefficient values at the 150th iteration. 
}
\label{fig:soa_analysis}
\end{figure}
Figure \ref{fig:soa_analysis}(a) illustrates the prices of individuals for SOA at the 150th iteration, considering various coefficient values. Specifically, we vary the coefficients for the fitness, represented as '$a$ $\times$ compatibility + $b$ $\times$ price', which in turn affects the resulting price values. In Figure \ref{fig:soa_analysis}(b), we show the compatibilities for the same individuals. In Figure \ref{fig:soa_analysis}(a) and \ref{fig:soa_analysis}(b), we mark the individual with the best price and compatibility using a red rectangle. Their coefficient values are the same, meaning that they are the same individual. For our equation, Fitness = $a$ $\times$ compatibility + $b$ $\times$ price, the coefficient values are $a$ = 0.957 and $b$ = 0.043 (keeping $a+b = 1$) for that individual. So, running SOA for different coefficient values, we are able to find a coefficient value that gives significantly better results than other coefficient values.

From actual land-use, Compatibility ($c$) = 3,330,225,944,839, Price ($p$) = 132,186,906,096.6. $c/p=25.19=a/b$ (we set), also $a+b=1$. Solving, we get $a=0.0382$ and $b=0.9619$, which are very close to the best coefficient values we get from our optimization runs using different coefficient values (Figure \ref{fig:soa_analysis}).

\begin{table}[!htbp]
\footnotesize
\centering
\caption{Parameters of the algorithms with relaxed constraints. Values in the green shaded cells are the same as the original.}
\begin{tabular}{lcc}
\toprule
Algorithm & \begin{tabular}{@{}l@{}}Area constraint, $\gamma$\\(initially 30\%)\end{tabular} & \begin{tabular}{@{}l@{}}Plot constraint, $\mu$\\(initially 20\%)\end{tabular} \\
\midrule
Mutation+SBX NSGA-II\_A & 40\% & \cellcolor{nice_green}20\% \\\cdashlinelr{1-3}
Mutation+SBX NSGA-II\_B & 60\% & \cellcolor{nice_green}20\% \\\cdashlinelr{1-3}
Mutation+SBX NSGA-II\_C & 80\% & \cellcolor{nice_green}20\% \\\cdashlinelr{1-3}
Mutation+SBX NSGA-II\_D & 100\% & \cellcolor{nice_green}20\% \\\cdashlinelr{1-3}
Mutation+SBX NSGA-II\_E & 40\% & 100\% \\\cdashlinelr{1-3}
CR+DES\_A & 40\% & \cellcolor{nice_green}20\% \\\cdashlinelr{1-3}
CR+DES\_B & 60\% & \cellcolor{nice_green}20\% \\\cdashlinelr{1-3}
CR+DES\_C & 80\% & \cellcolor{nice_green}20\% \\\cdashlinelr{1-3}
CR+DES\_D & 100\% & \cellcolor{nice_green}20\% \\\cdashlinelr{1-3}
CR+DES\_E & 40\% & 100\% \\\cdashlinelr{1-3}
CR+DES\_F & 60\% & 100\% \\\cdashlinelr{1-3}
CR+DES\_G & 80\% & 100\% \\\cdashlinelr{1-3}
CR+DES\_H & 200\% & 20\% \\\cdashlinelr{1-3}
SOA\_A & 40\% & 100\% \\
\bottomrule
\end{tabular}
\label{tab:relaxed_constraints}
\end{table}

\subsection{Constraint Relaxation}
We relax some of the constraints during optimization to explore the search space further, but at the final iteration, they are unrelaxed again so that we may get valid solutions only. This allows for more exploration within the search space. Relaxing the $\gamma$ constraint allows the optimization process to change the area of a land-use more, while relaxing $\mu$ constraint allows more plots to get their land-use changed. We don't relax constraints $P_{\text{min}}$ and $P_{\text{max}}$ to prevent the optimization process from uncontrollably exploring the search space. In Table \ref{tab:relaxed_constraints}, we have shown the changed values of the constraints and the corresponding method names.

\subsubsection{Results of Constraint Relaxation}
\begin{figure}[h]
\centering
\includegraphics[width=1\textwidth]{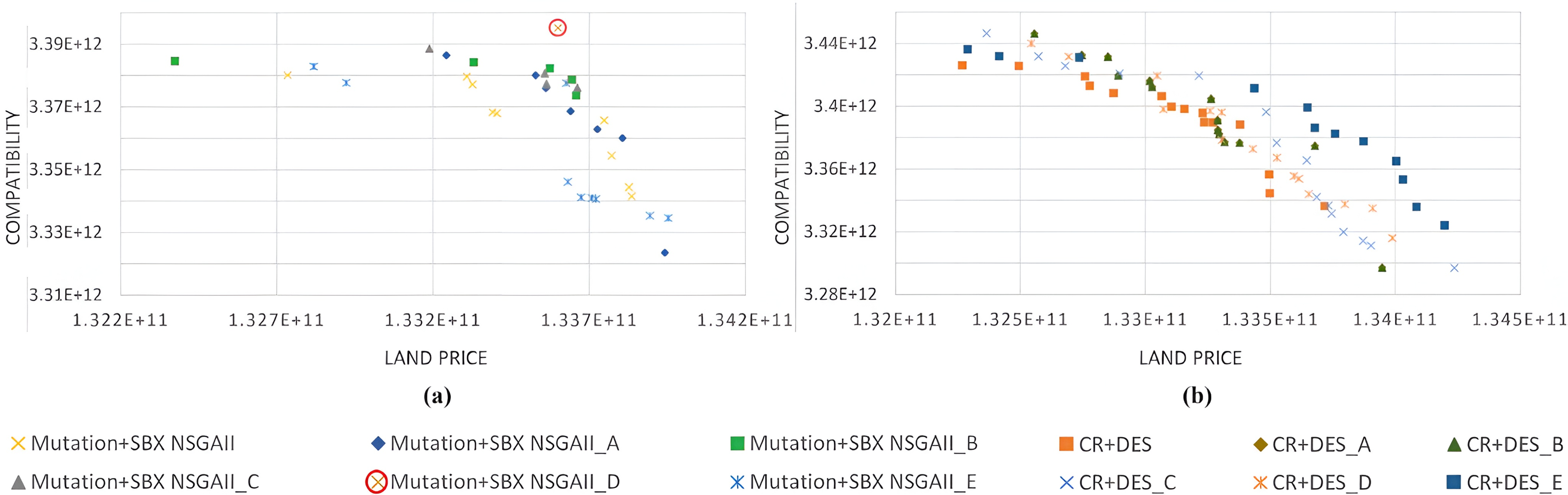}
\caption{(a) Pareto Fronts of original Mutation+SBX NSGA-II and relaxed constraint versions of the method (after 150 iterations). 
(b) Pareto Fronts of original CR+DES and relaxed constraint versions of the method (after 150 iterations). 
}
\label{fig:relaxed_constraints}
\end{figure}

Figure \ref{fig:relaxed_constraints}(a) showcases the Pareto Fronts of Mutation+SBX NSGA-II methods with relaxed constraints (as outlined in Table \ref{tab:relaxed_constraints}). During the optimization process, the constraint $\gamma$ (and $\mu$ in some cases) is relaxed, but at the final iteration (150th iteration), the $\gamma$ constraint is unrelaxed. Even though the values of constraints are relaxed during optimization, there are many solutions after the 150th iteration that satisfy the initial unrelaxed constraints.

When only $\gamma$ constraint is relaxed (Mutation+SBX NSGA-II\_A, Mutation+SBX NSGA-II\_B, Mutation+SBX NSGA-II\_C, and Mutation+SBX NSGA-II\_D), no solution that dominates all the solutions of the unrelaxed method is generated, but improvement of compatibility (for a similar price in a price range) from unrelaxed Mutation+SBX NSGA-II is observed. When both $\gamma$ and $\mu$ constraints are relaxed (Mutation+SBX NSGA-II\_E), no solution that dominates all the solutions of the unrelaxed method is generated. But the solutions are much more spread out. In terms of the best compatibility, Mutation+SBX NSGA-II is better, but in terms of the best price, Mutation+SBX NSGA-II\_E is better (Figure \ref{fig:relaxed_constraints}(a)). Also, further visual inspection confirms that all of the Pareto Fronts are overlapping except for that of Mutation+SBX NSGA-II\_D. Also, by aggregating all non-dominated solutions from Pareto Fronts and filtering out dominated ones, we may obtain a global Pareto Front of all solutions. By visual inspection, we see that such a Pareto Front will have solutions from all of the methods (Figure \ref{fig:relaxed_constraints}). This indicates that we can run multiple methods in parallel to get better solutions.

Figure \ref{fig:relaxed_constraints}(b) showcases the Pareto Fronts of CR+DES methods with relaxed constraints (as outlined in Table \ref{tab:relaxed_constraints}). Recall that, during the optimization process, the constraint $\gamma$ (and $\mu$ in some cases) is relaxed, but at the final iteration (150th iteration), the $\gamma$ constraint is unrelaxed. Even though the values of constraints are relaxed during optimization, there are many solutions after the 150th iteration that satisfy the initial unrelaxed constraints. Inspecting closely, we see that solutions from CR+DES\_C dominate all solutions from CR+DES, indicating CR+DES\_C to be a clear winner when comparing with CR+DES. Also, the combined Pareto Front of relaxed CR+DES is better and also clearly outperforms CR+DES.

When only $\gamma$ constraint is relaxed (CR+DES\_A, CR+DES\_B, CR+DES\_C, and CR+DES\_D), improvement of price (for similar compatibility) from unrelaxed method CR+DES (Figure \ref{fig:relaxed_constraints}(b)) is observed. When both $\gamma$ and $\mu$ constraints are relaxed (CR+DES\_E), no solution that dominates all the solutions of the unrelaxed method is generated. But improvement of price (for similar compatibility) is observed (Figure \ref{fig:relaxed_constraints}(b)). We also do optimization with similar relaxed constrained versions of more variants. The performance do not improve much in those cases. Please refer to Supplementary Sections S3.3.1, S3.3.2, S3.3.3, and S3.3.4 and Supplementary Figures S2-12.

\subsubsection{Land-use change in solutions with relaxed constraints}
In this subsection, we look at all the individuals of the 150th iteration for relaxed constraints versions of Mutation+SBX NSGA-II (Mutation+SBX NSGA-II\_D) and CR+DES (CR+DES\_D). This allows us to see how many of the solutions are well within the original constraints (defined in Table \ref{tab:parameters}, other than the yellow shaded constraint). Also, we will be able to see what solutions are not within but closer to the original constraints.

\begin{figure}[h]
\centering
\includegraphics[width=1\textwidth]{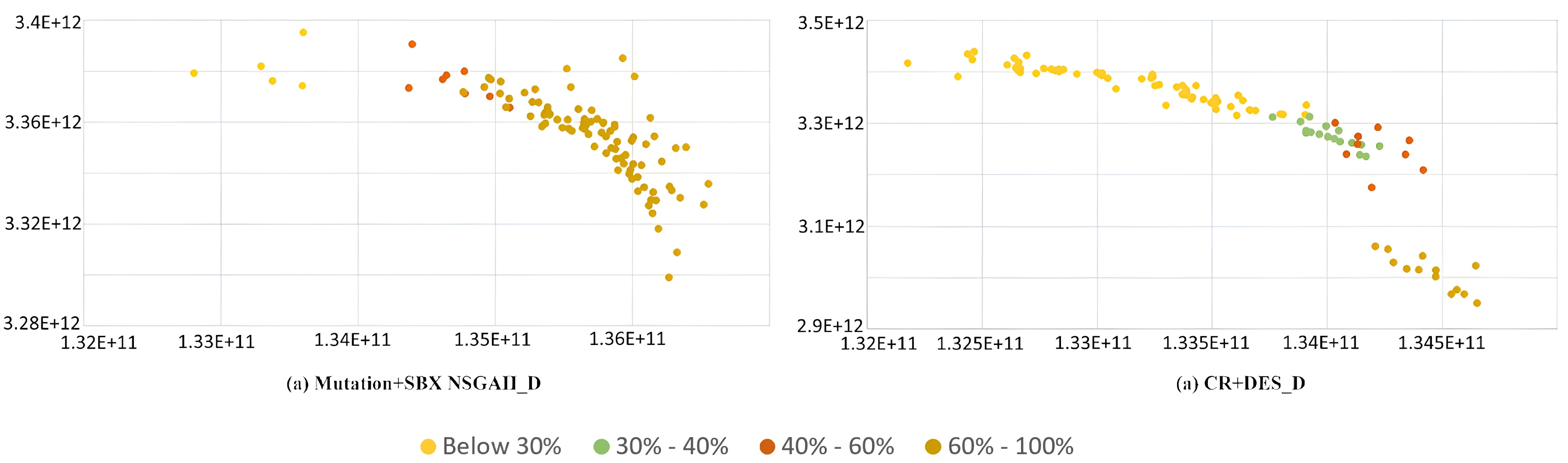}
\caption{Solutions of Mutation+SBX NSGA-II\_D and CR+DES\_D after 150th Iteration. The change in area (for the land-use with the highest area change) for a solution is indicated by its color.}
\label{fig:landuse_change}
\end{figure}
Figure \ref{fig:landuse_change}(a) presents the solutions of Mutation+SBX NSGA-II\_D 150th Iteration. A total of 5 solutions satisfy the initial 30\% Land-use change constraint. For solutions with more land-use change, compatibility tends to be worse, but price increases. Figure \ref{fig:landuse_change}(b) presents the solutions of CR+DES\_D after 150th Iteration. A total of 61 solutions satisfy the initial 30\% Land-use change constraint. In this case, also, for solutions with more land-use change, compatibility tends to be worse, but price increases.

\subsubsection{Relaxing Constraints Further} \label{sec:f}
We further relax the constraints for one of the well-performing methods, CR+DES, and observe the outcome. Additionally, we investigate whether the results continue to improve as we progressively relax the constraints, or if they remain consistent or potentially worsen. The price constraints are the same throughout the runs, but the land-use constraints are changed. We run the algorithms for 150 iterations.

\begin{figure}[h]
\centering
\includegraphics[width=0.65\textwidth]{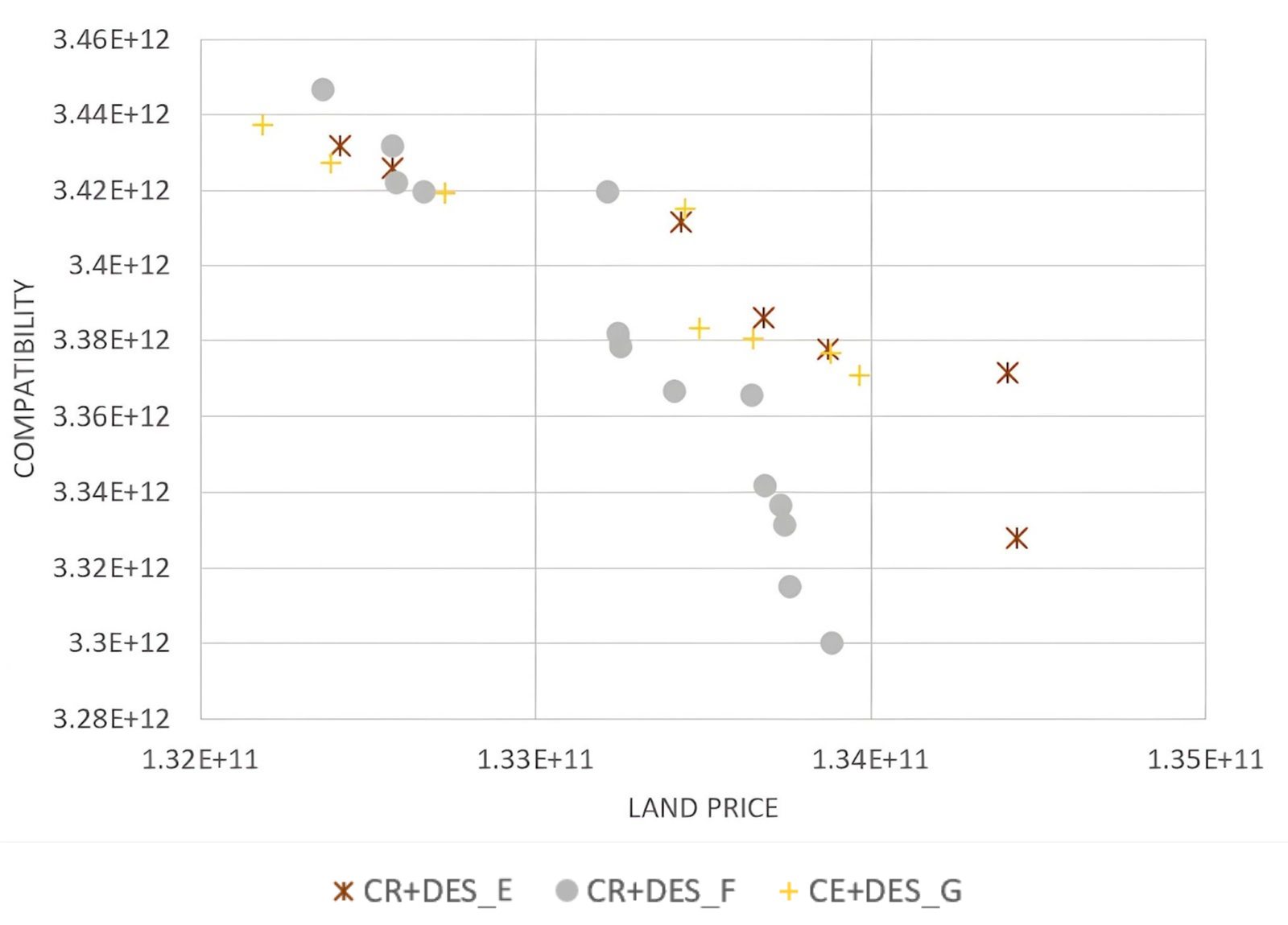}
\caption{Effects of relaxing area constraint and also relaxing plot change constraint.}
\label{fig:further_relaxation}
\end{figure}

In Figure \ref{fig:further_relaxation}, we can observe the impact of relaxing the area constraint as well as the plot change constraint. As we progressively relax these constraints, the best prices decrease. If we look at compatibility, for CR+DES\_F, there is a solution with the best compatibility. As we relax constraints more, the best compatibility decreases, as does the HV value. Looking at the figure, we also see that the Pareto Front of further relaxed methods (CR+DES\_F and CR+DES\_G) overlaps with the less relaxed methods (CR+DES\_E). Also, the combined Pareto Front of further relaxed runs overlaps with less relaxed methods, indicating that too much relaxation of constraints may have adverse effects.

All of our algorithms encounter a state of stagnation when we allow 100\% plots to be changed by the genetic algorithm, possibly due to getting trapped in a local minima. To solve this problem, we allow any area change during optimization. But it gives a solution set where no solution fulfills the initial set of constraints. The impact of this approach is evaluated across CR+DES and SOA (a similar analysis for more methods is presented in Supplementary Sections S1 and S2).

\subsubsection{Analysis with no constraint (NC)}
In this setting, we adopt a flexible approach that does not impose any constraints on land changes, thereby permitting any type of land alteration. Additionally, we do not impose restrictions on price changes, allowing them to vary freely. This flexibility in both land-use changes and prices ensures a comprehensive exploration of the possibilities without imposing predefined limitations or restrictions on the outcomes. But this analysis do not consider the constraints, even when we have solutions at the final iteration. So, it breaks the optimization constraints specified in Section \ref{sec:cn}. So, the final result may not be of much use. Nevertheless, the analysis is for getting an idea of where the methods end up in the search space. CR+DES produces a diverse set of solutions in the objective space, indicating broad exploration across the search space. In contrast, methods such as Mutation+SBX NSGA-II produce solutions that are more tightly clustered in the objective space, suggesting convergence possibly toward a local or global optimum in the search space (please refer to Supplementary Section S3.1 and Supplementary Figure S1).

\subsection{Convergence Analysis}
To check whether our algorithms are performing as intended, we look at Pareto Fronts of different generations. Looking at them allows us to see whether there are noticeable improvements of prices and compatibility, or our algorithms are stagnating. Also, looking at convergence allows us to assume whether more improvement is possible by running more iterations. We look at the convergence of CR+DES\_A (due to CR+DES being one of the best performing algorithms).

\begin{figure}[!htbp]
\centering
\includegraphics[width=0.95\textwidth]{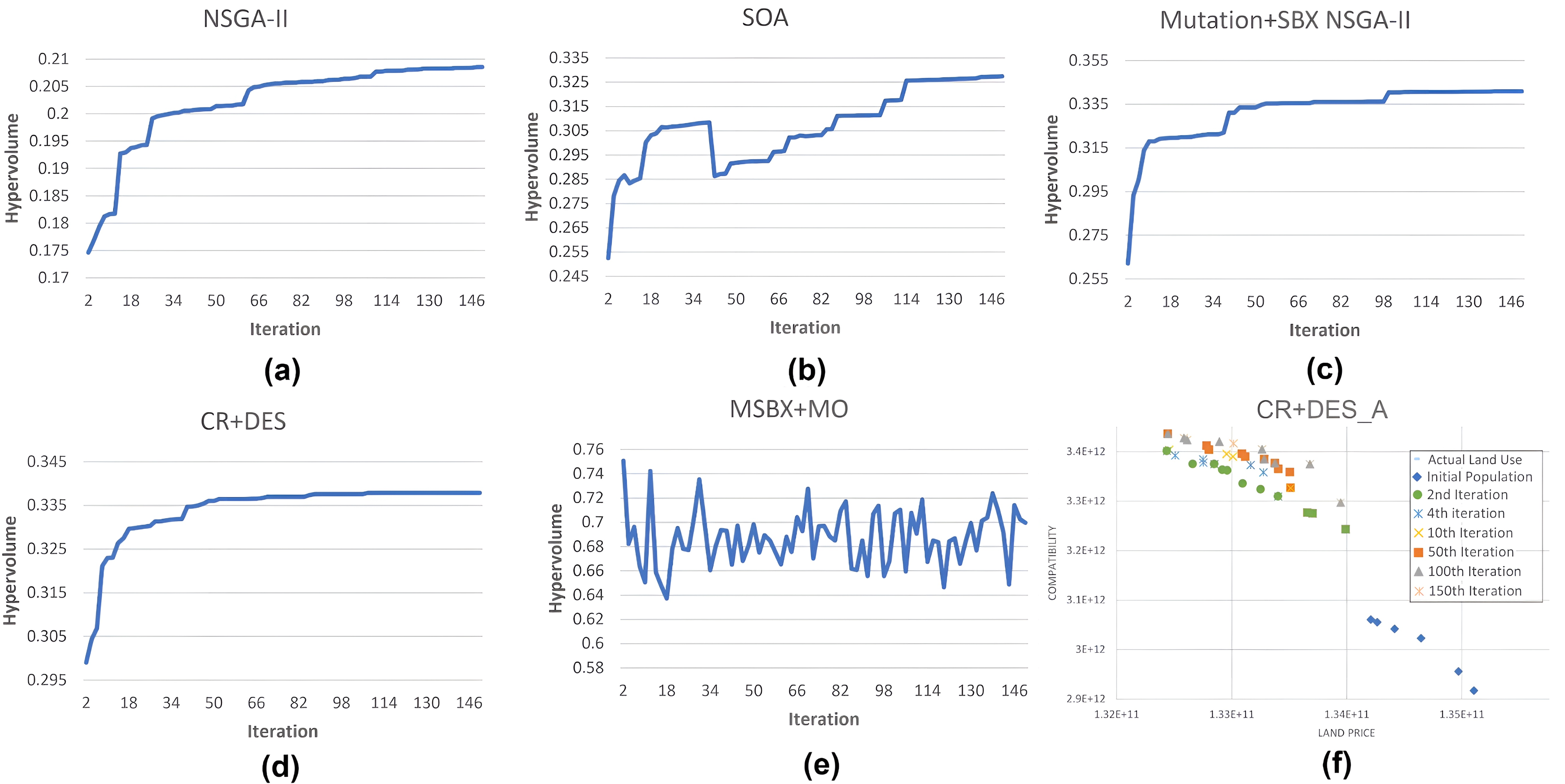}
\caption{(a)-(e) shows average HV values of five runs for NSGA-II, SOA, Mutation+SBX NSGA-II, CR+DES, and MSBX+MO for 2-150th iteration. (f) Shows the actual land-use and the progression of the Pareto front for CR+DES\_A at the initial population and the 2nd, 4th, 10th, 50th, 100th, and 150th iterations.
}
\label{fig:convergence_analysis}
\end{figure}
Figure \ref{fig:convergence_analysis}(a)–\ref{fig:convergence_analysis}(e) presents the average hypervolume (HV) values from five runs of NSGA-II, SOA, Mutation+SBX NSGA-II, CR+DES, and MSBX+MO, covering iterations 2 through 150. NSGA-II (Figure \ref{fig:convergence_analysis}(a)), Mutation+SBX NSGA-II (Figure \ref{fig:convergence_analysis}(c)), and CR+DES (Figure \ref{fig:convergence_analysis}(d)) demonstrate a gradual increase in HV values leading to convergence. In contrast, SOA (Figure \ref{fig:convergence_analysis}(b)) initially shows a decrease in HV values before converging at a higher level, while the HV values for MSBX+MO (Figure \ref{fig:convergence_analysis}(e)) do not converge, but they are at a higher value.

In Figure \ref{fig:convergence_analysis}(f), in the initial population, the individuals exhibit noticeably lower compatibility but offers higher prices. However, in the very next generation, we witness a dramatic surge in compatibility accompanied by a significant decrease in prices. As we progress through subsequent generations, the prices remains relatively stable, with minimal fluctuation, while compatibility continues to steadily improve. Also, between the 100th and 150th iterations, no discernible changes are evident, indicating that the algorithm indeed converges. We also analyze convergence for CR+DES\_B variants. The algorithms seems to converge (Please refer to Supplementary Section S5 and Supplementary Figure S16).

\subsection{Land-use distribution}
\begin{figure}[!htbp]
\centering
\includegraphics[width=1\textwidth]{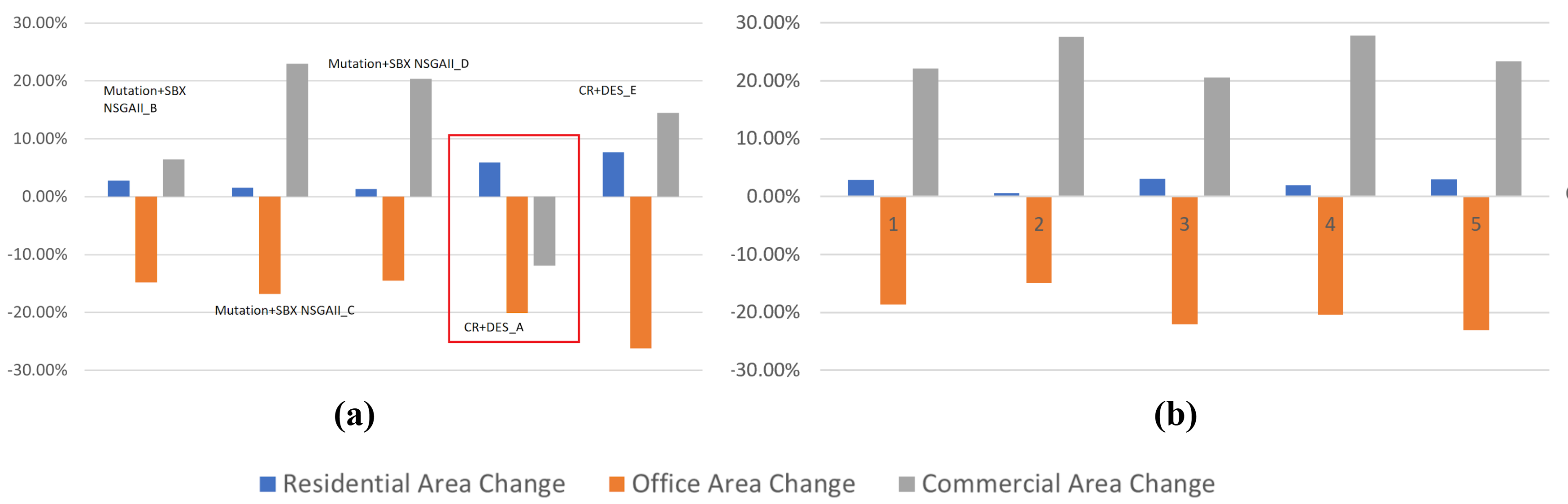}
\caption{Land-use change for better performing optimization algorithm. (a) For the individual with the best compatibility (running 150th iterations), the land-use change of three types (residential, office, and commercial) based on area is shown. (b) For the individual with the best price (running 150 iterations), the land-use change of three types (residential, office, and commercial) based on area is shown.
}
\label{fig:landuse_distribution}
\end{figure}

We also look at the land-use distribution of our final results generated by optimization algorithms that have solutions in the combined Pareto Front of all algorithms. The algorithms are, Mutation+SBX NSGA-II\_B, Mutation+SBX NSGA-II\_C, Mutation+SBX NSGA-II\_D, CR+DES\_A and CR+DES\_B. For the algorithms with the best results, the land-use change of three types (residential, office, and commercial) based on the total floor area they encompass is shown in Figure \ref{fig:landuse_distribution}. A huge decrease in the commercial area and an increase in the office area can be noticed. For CR+DES\_A, both office and commercial spaces decrease. Both CR+DES\_A and CR+DES\_E has the highest increase in residential area. For individuals with the best prices, a consistent increase in office area and decrease in commercial area is seen (irrespective of the five algorithms reported in Figure \ref{fig:landuse_distribution}). Please refer to Supplementary Section S6 and Supplementary Figures S22-S26 for details on how each method changes land-use and land-use area.

\subsection{Comparison with the state of the art}
We compare Pareto Fronts and Pareto optimal points of our multi Objective approaches and the best solution for our SOA approach with state of the art, NSGA-II method \cite{Sharmin2019} for our study area. Examining the collection of all results combined allows us to identify the better performing methods. Also, we are able to see how the relaxed constrained versions perform.

\subsubsection{Comparing Pareto Fronts}
In Table \ref{tab:performance_metrics}, we present the HV, GD, GD+, IGD, and IGD+ values for the Pareto fronts obtained from different approaches (price and compatibility are normalized before calculating the values). For calculating HV, the reference vector is $v=(0,0)$. Each optimization algorithm is executed five times, and the reported metrics are the averages across these runs. The table contains the results of 20 algorithms. The green shaded cells of the table contain the highest values for an indicator. Higher HV value indicates better performance, whereas lower GD, GD+, IGD, and IGD+ values indicate better performance. Looking at Table \ref{tab:performance_metrics}, we notice that the highest HV is achieved by MSBX+MO. Also, the lowest GD and GD+ values are achieved by MSBX+MO. The lowest IGD and IGD+ values are achieved by CR\_DES\_E. Indicating their superiority over other methods.

{\renewcommand{\arraystretch}{1.1}
\begin{table}[h]
\footnotesize
\centering
\caption{Hypervolume (HV), Generational Distance (GD), Generational Distance Plus (GD+), Inverted Generational Distance (IGD), and Inverted Generational Distance Plus (IGD+) value of Pareto Fronts.}
\begin{tabular}{p{6cm}p{1.2cm}p{1.2cm}p{1.2cm}p{1.2cm}p{1.2cm}}
\toprule
Method & HV & GD & GD+ & IGD & IGD+ \\
\midrule
NSGA-II & 0.162 & 0.4223 & 0.4223 & 0.4266 & 0.4266 \\ 
\cdashlinelr{1-6}
CR+DES & 0.4165 & 0.1171 & 0.117 & 0.1732 & 0.1719 \\
\cdashlinelr{1-6}
CR+DES\_A & 0.4682 & 0.0927 & 0.0924 & 0.1366 & 0.1342 \\
\cdashlinelr{1-6}
CR+DES\_B & 0.4682 & 0.0927 & 0.0924 & 0.1366 & 0.1342 \\
\cdashlinelr{1-6}
CR+DES\_C & 0.4858 & 0.1368 & 0.1367 & 0.1405 & 0.1353 \\
\cdashlinelr{1-6}
CR+DES\_D & 0.4769 & 0.1046 & 0.1043 & \cellcolor{nice_green}0.1464 & \cellcolor{nice_green}0.1445 \\
\cdashlinelr{1-6}
CR+DES\_E & 0.5543 & 0.0661 & 0.0654 & 0.1088 & 0.1013 \\
\cdashlinelr{1-6}
CR+DES\_F & 0.5383 & 0.0788 & 0.0782 & 0.1161 & 0.113 \\
\cdashlinelr{1-6}
CR+DES\_G & 0.5077 & 0.0842 & 0.084 & 0.1314 & 0.1292 \\
\cdashlinelr{1-6}
CR+DES\_H & 0.5011 & 0.099 & 0.0989 & 0.1238 & 0.122 \\
\cdashlinelr{1-6}
MSBX\_MO & \cellcolor{nice_green}0.7872 & \cellcolor{nice_green}0.0426 & \cellcolor{nice_green}0.0145 & 0.3702 & 0.1907 \\
\cdashlinelr{1-6}
Mutation+SBX NSGA-II & 0.3959 & 0.13 & 0.13 & 0.2254 & 0.2167 \\
\cdashlinelr{1-6}
Mutation+SBX NSGA-II\_A & 0.4034 & 0.1192 & 0.1192 & 0.209 & 0.2049 \\
\cdashlinelr{1-6}
Mutation+SBX NSGA-II\_B & 0.3797 & 0.0898 & 0.0896 & 0.2243 & 0.2084 \\
\cdashlinelr{1-6}
Mutation+SBX NSGA-II\_C & 0.3596 & 0.0946 & 0.0943 & 0.2417 & 0.2206 \\
\cdashlinelr{1-6}
Mutation+SBX NSGA-II\_D & 0.3722 & 0.0611 & 0.0611 & 0.2409 & 0.202 \\
\cdashlinelr{1-6}
Mutation+SBX NSGA-II\_E & 0.3917 & 0.1366 & 0.1366 & 0.2464 & 0.2317 \\
\cdashlinelr{1-6}
SOA & 0.2858 & 0.1249 & 0.1249 & 0.2719 & 0.2563 \\
\cdashlinelr{1-6}
SOA\_A & 0.3291 & 0.0878 & 0.0878 & 0.2863 & 0.2612 \\
\cdashlinelr{1-6}
TwoArc & 0.1241 & 0.4539 & 0.4539 & 0.4905 & 0.4905 \\
\bottomrule
\end{tabular}
\label{tab:performance_metrics}
\end{table}
}

\begin{figure}[!htbp]
\centering
\includegraphics[width=0.95\textwidth]{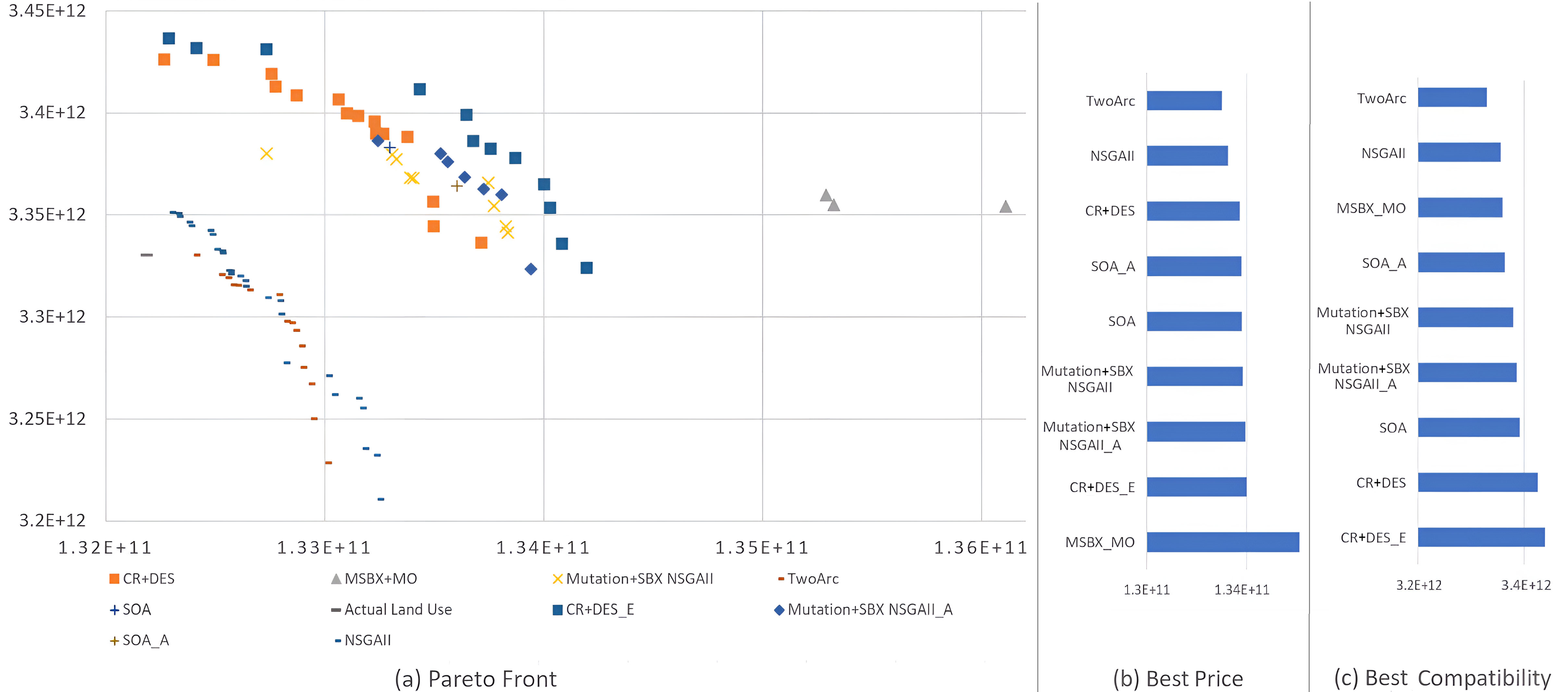}
\caption{(a) Comparison of optimization algorithms and their relaxed version with best IGD+, with the state-of-the art approach NSGA-II method \cite{Sharmin2019}. Each method is run 5 times, and the combined Pareto front for each method is plotted. (b) A bar chart depicting the best price for each of our optimization algorithms, their relaxed versions with best IGD+ and state-of-the-are NSGA-II, from 5 runs. (c) A bar chart depicting the best compatibility for each of our optimization algorithms, their relaxed versions with best IGD+ and state-of-the-are NSGA-II, from 5 runs.}
\label{fig:sota_comparison}
\end{figure}
Looking at Figure \ref{fig:sota_comparison}(a), most of our proposed metaheuristic optimization algorithms perform visually better than state-of-the-art NSGA-II. Looking at the combined Pareto Front, the Metaheuristic optimization algorithms with solutions in that Pareto Front are CR+DES\_E and MSBX+MO (Figure \ref{fig:sota_comparison}(a)). Also, visually, CR+DES methods are much more spread out, indicating better performance. Also, solutions from CR+DES\_E dominate all solutions from NSGA-II. Looking at Figure \ref{fig:sota_comparison}(b), the best price is achieved by MSBX+MO, and from Figure \ref{fig:sota_comparison}(c), the best compatibility is achieved by CR+DES\_E. More comparison results are in the Supplementary File (Section S5).
\subsubsection{Comparing Pareto optimal points}
We can categorize the final solution as Type I, Type II, and Type III as follows. Type I refers to the solutions with the best compatibility; Type II refers to the solutions with the best land price, and Type III refers to the solutions with the highest crowding distance (average distance between a solution and its two nearest neighbors in objective space). So, Type I and Type II solutions represent the Pareto optimal points, while Type III solutions, although not Pareto optimal, aid in contrasting the overall solution. In Table \ref{tab:pareto_optimal_comparison} we present the compatibility and price of Types I, II, and III solutions for some of our better performing algorithms. We also show by what percentage the compatibility or price changes from the price and compatibility that was seen in actual land-use allocation (i.e., the use allocation that was in the dataset).

Table \ref{tab:pareto_optimal_comparison} shows that CR+DES\_A has the overall best compatibility with competitive prices. MSBX+MO gives the best prices for Type I and the best compatibility for Type II. Also, Among the optimization strategies employed, MSBX+MO stands out for its distinct approach, employing mutation before crossover, rendering it simpler than some counterparts. However, its simplicity still has its effectiveness. It yielded favorable prices, as is evident in Table \ref{tab:pareto_optimal_comparison}. Furthermore, the Type II compatibility of MSBX+MO surpassed all other approaches. Mutation+SBX NSGA-II\_D has the best Type II price and overall price (3.3\% increase in price). CR+DES\_B has the best compatibility for Type III, and Mutation+SBX NSGA-II\_D has the best price. Compared to the state-of-the-art, we see noticeable improvements.
{\renewcommand{\arraystretch}{1.1}
\begin{table}[h]
\footnotesize
\centering
\caption{Comparison of the best results (values divided by $10^{12}$). The percentage indicates the increase from the actual land‐use map, where compatibility is 3.3302 and price is 0.1322.}
\begin{tabular}{p{5.5cm}
                p{1.2cm} p{1.2cm}
                p{1.2cm} p{1.2cm}
                p{1.2cm} p{1.2cm}}
\toprule
Approach  
  & \begin{tabular}{@{}l@{}}Type\\I\\Comp.\end{tabular}  
  & \begin{tabular}{@{}l@{}}Type\\I\\Price\end{tabular}  
  & \begin{tabular}{@{}l@{}}Type\\II\\Comp.\end{tabular}  
  & \begin{tabular}{@{}l@{}}Type\\II\\Price\end{tabular}  
  & \begin{tabular}{@{}l@{}}Type\\III\\Comp.\end{tabular}  
  & \begin{tabular}{@{}l@{}}Type\\III\\Price\end{tabular} \\
\midrule
NSGA-II \cite{Sharmin2019}  
  & 3.3562       & 0.1323 
  & 3.2322       & 0.1332 
  & 3.2774       & 0.1328 \\
\cdashlinelr{1-7}
Mutation+SBX NSGA-II  
  & 3.3846       & 0.1331 
  & 3.3451       & 0.1338 
  & 3.3628       & 0.1338 \\
\cdashlinelr{1-7}
Mutation+SBX NSGA-II\_D  
  & 3.3952       & 0.1336 
  & 3.3358 
    & \cellcolor{nice_green}\begin{tabular}{@{}l@{}}0.1366\\(3.30\%)\end{tabular} 
    & 3.3851 
    & \cellcolor{nice_green}\begin{tabular}{@{}l@{}}0.1359\\(2.83\%)\end{tabular} \\
\cdashlinelr{1-7}
MSBX+MO  
  & 3.3542 
    & \cellcolor{nice_green}\begin{tabular}{@{}l@{}}0.1350\\(2.16\%)\end{tabular} 
  & \cellcolor{nice_green}\begin{tabular}{@{}l@{}}3.3531\\(0.69\%)\end{tabular}
    & 0.1359  
    & N/A      & N/A \\
\cdashlinelr{1-7}
SOA  
  & 3.3923       & 0.1329 
  & N/A          & N/A 
  & N/A          & N/A \\
\cdashlinelr{1-7}
CR+DES  
  & 3.4240       & 0.1321 
  & 3.2737       & 0.1337 
  & 3.3444       & 0.1335 \\
\cdashlinelr{1-7}
CR+DES\_A  
  & \cellcolor{nice_green}\begin{tabular}{@{}l@{}}3.4356\\(3.16\%)\end{tabular} 
    & 0.1324 
  & 3.2970       & 0.1339 
  & 3.3748       & 0.1337 \\
\cdashlinelr{1-7}
CR+DES\_B  
  & 3.4317       & 0.1324 
  & 3.3279       & 0.1344 
  & \cellcolor{nice_green}\begin{tabular}{@{}l@{}}3.4115\\(2.44\%)\end{tabular}
    & 0.1334 \\

\bottomrule
\end{tabular}
\label{tab:pareto_optimal_comparison}
\end{table}
}
\subsection{Statistical Analysis}
\begin{figure}[!htbp]
  \centering

  \begin{subfigure}[b]{0.48\textwidth}
    \centering
    \includegraphics[width=\linewidth]{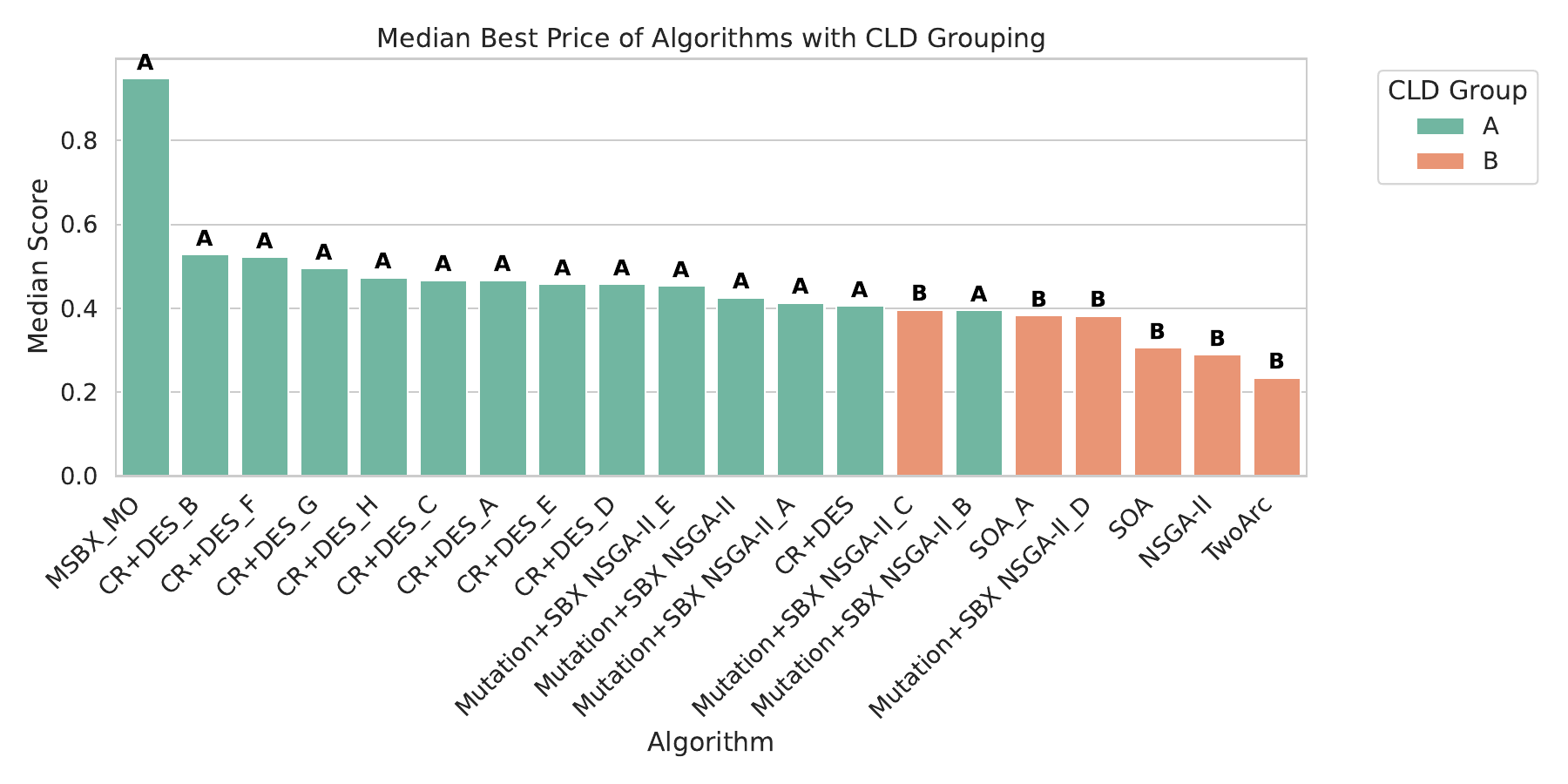}
    \caption{}
    \label{fig:price_cld}
  \end{subfigure}
  \hfill
  \begin{subfigure}[b]{0.48\textwidth}
    \centering
    \includegraphics[width=\linewidth]{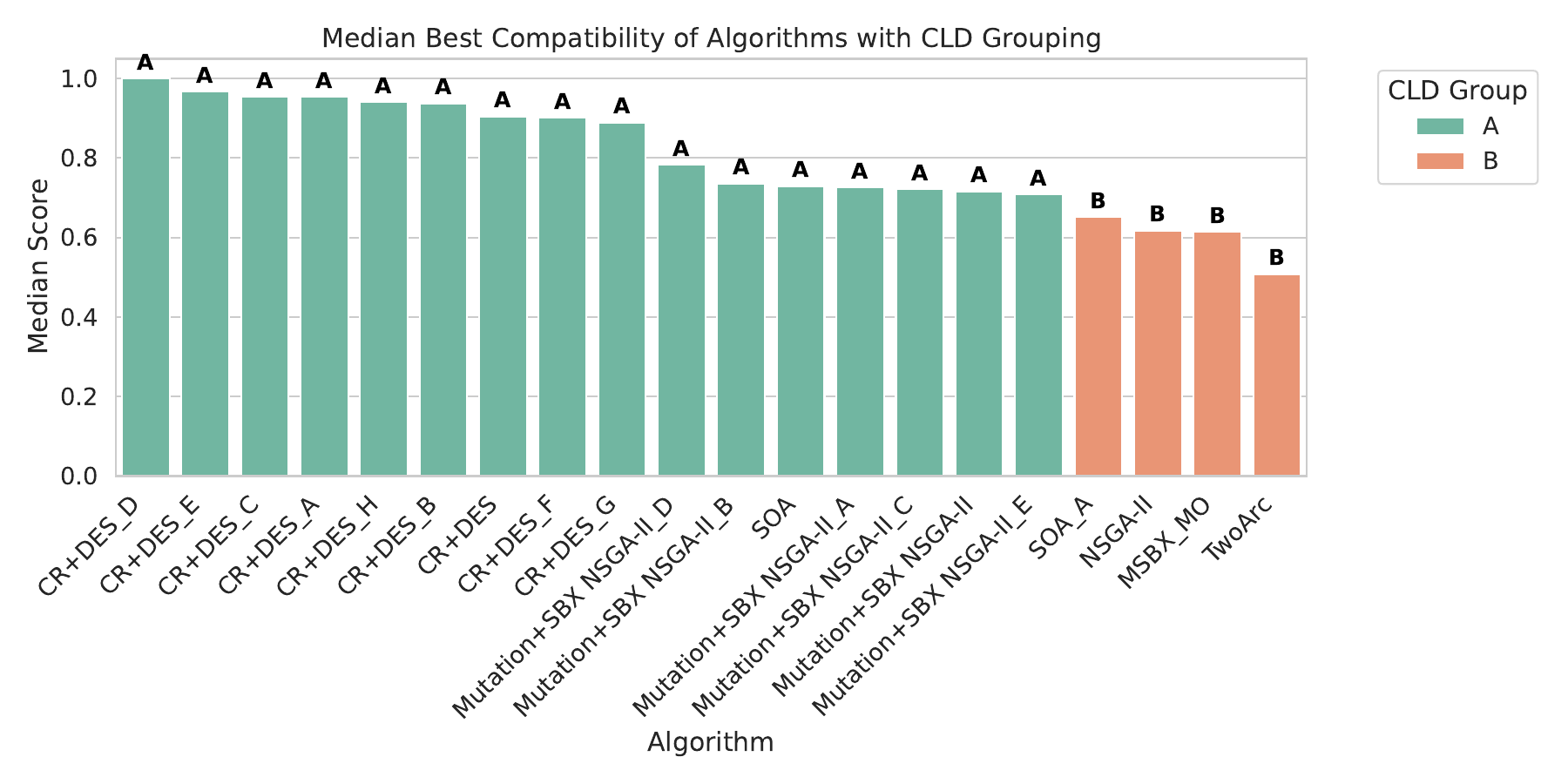}
    \caption{}
    \label{fig:compat_cld}
  \end{subfigure}

  \caption{(a) Median Best Prices of Algorithms with CLD groupings. (b) Median Best Compatibilities of Algorithms with CLD groupings.}
  \label{fig:Price_comp}
\end{figure}

\begin{figure*}[!htbp]
  \centering

  \begin{subfigure}[b]{0.48\textwidth}
    \centering
    \includegraphics[width=\linewidth]{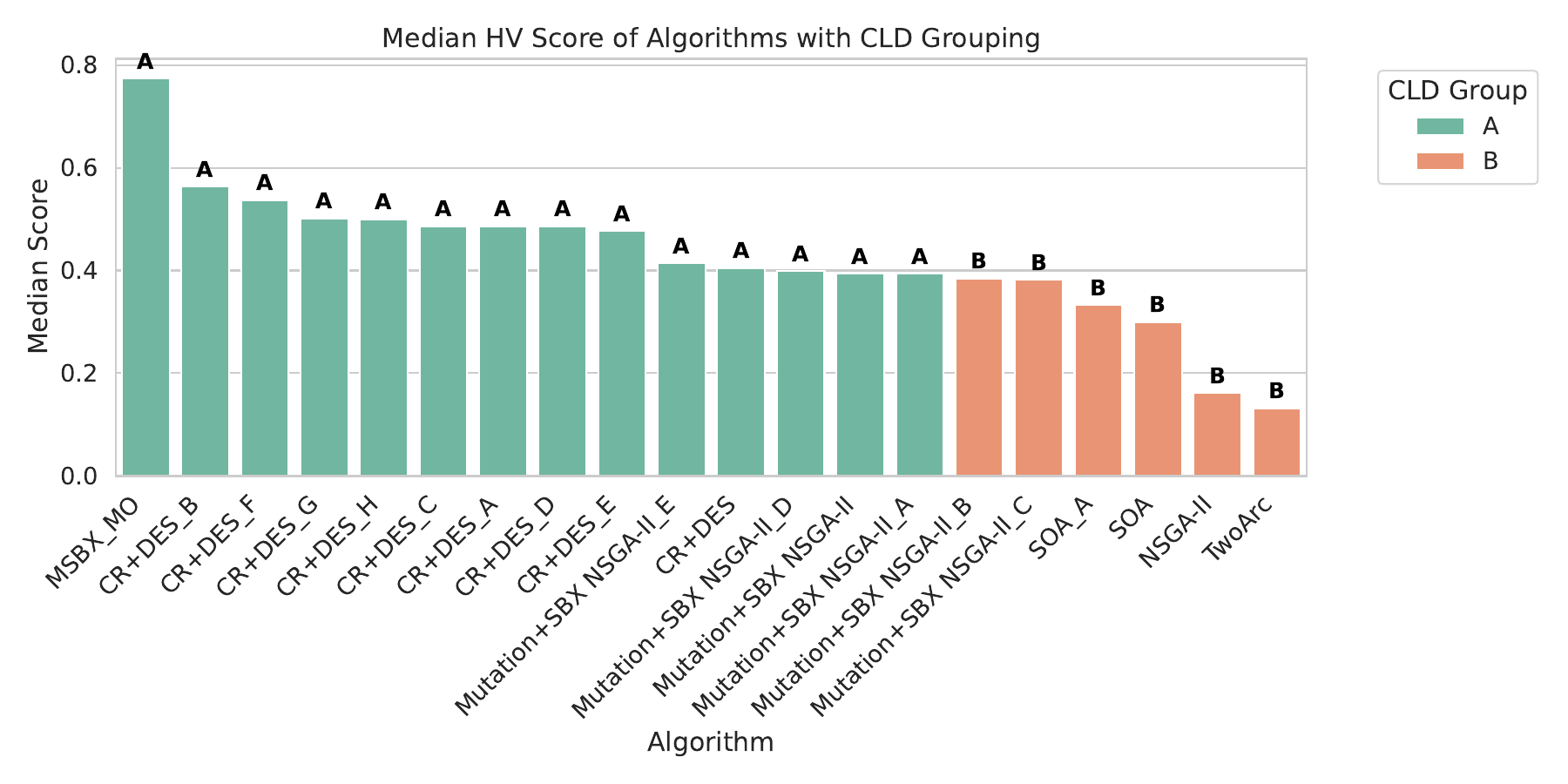}
    \caption{}
    \label{fig:hv_cld}
  \end{subfigure}
  \hfill
  \begin{subfigure}[b]{0.48\textwidth}
    \centering
    \includegraphics[width=\linewidth]{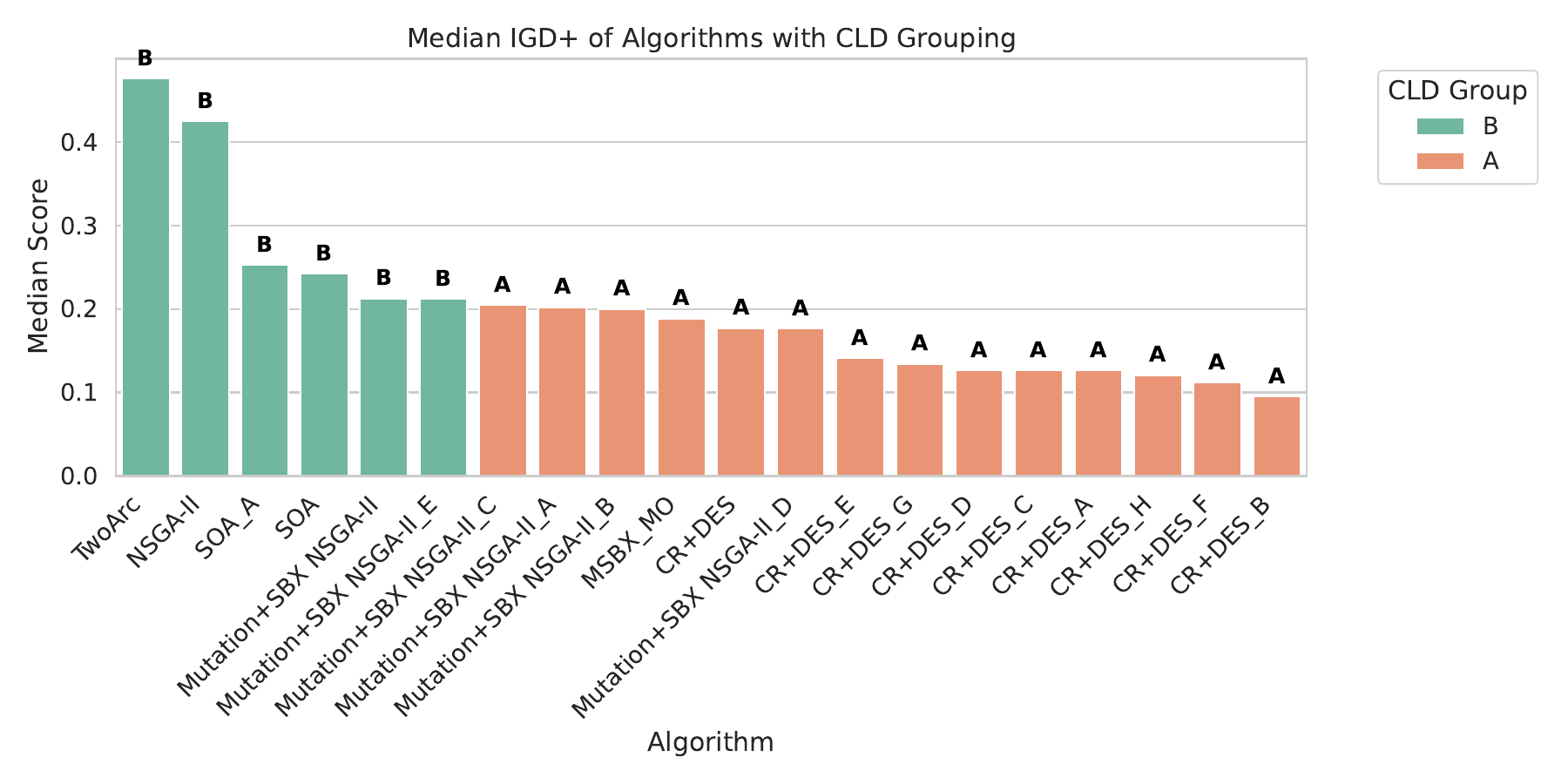}
    \caption{}
    \label{fig:igd_cld}
  \end{subfigure}

  \caption{(a) Median HV Scores of Algorithms with CLD groupings. (b) Median IGD\(+\) Scores of Algorithms with CLD groupings.}
  \label{fig:HV_IGD}
\end{figure*}

We compared 20 algorithms across four metrics—best price, best compatibility, hypervolume (HV), and IGD+—using the Kruskal–Wallis test. In all cases, the resulting $p$-values were $< 0.05$, indicating significant differences among the algorithms. We therefore performed Bonferroni-corrected Dunn’s post-hoc tests to identify which pairs differed, summarizing the results with compact letter displays (CLDs) in Figure~\ref{fig:Price_comp} and Figure~\ref{fig:HV_IGD}. In a CLD, algorithms sharing a letter are not significantly different, whereas different letters denote significant differences.

Figure~\ref{fig:Price_comp}(a) shows that CR+DES and its relaxed variants form group~A (the higher-performing group for best price), as does MSBX+MO. CR+DES uses difference vectors, whereas none of the group~B methods do; this suggests that incorporating difference vectors in land-use optimization increases the best price. The relaxed versions of Mutation+SBX~NSGA-II fall into group~B, implying that excessive constraint relaxation reduces price performance. In Figure~\ref{fig:Price_comp}(b), the same pattern is seen for the CR+DES methods, but MSBX+MO now appears in group~B, signalling lower compatibility.

Figures~\ref{fig:HV_IGD}(a) and~\ref{fig:HV_IGD}(b) summarize the HV and IGD+ results. In both cases, CR+DES and its relaxed variants remain in group~A along with MSBX+MO, indicating superior performance on these metrics. Overall, CR+DES and MSBX+MO perform well on best price, HV, and IGD+, but MSBX+MO falls behind on compatibility, while CR+DES remains consistently strong.

\section{Discussion} \label{sec:d}
Based on our rigorous empirical evaluation, the CR+DES approach demonstrates the best performance based on compatibility, competitive performance based on price, and smooth convergence, as evident from the outcomes illustrated in our results. Statistical analysis using Kruskal-Wallis tests and Bonferroni-corrected Dunn's post-hoc comparisons confirms that CR+DES consistently belongs to the superior statistical group across four performance metrics (price, compatibility, HV, and IGD+). This notable improvement can be attributed to the incorporation of difference vectors alongside scaling in the candidate solutions list, a key insight supported by our statistical findings that algorithms utilizing difference vectors (CR+DES and MSBX+MO) can outperform those that do not. Also, when we relax constraints, CR+DES and Mutation+SBX NSGA-II produce individuals with better compatibility for a similar price or better price for similar compatibility than the individuals of their original constraint versions. This actually supports our hypothesis that the relaxed constrained versions of the algorithms are allowing further and better exploration of the search space. This is further evident from the HV values reported: the HV values of relaxed constrained versions of the algorithm are higher than those of the original constrained versions, indicating the effectiveness of relaxing constraints. However, it is important to note that relaxing the constraints too much can be detrimental. Allowing land-use change without any restriction at all does not even generate solutions that satisfy area constraints (please refer to Section \ref{sec:f}). Our statistical analysis reveals an important trade-off: while MSBX+MO performs exceptionally well for price optimization and falls into the superior statistical group for HV and IGD+ metrics, it shows weaker performance in compatibility compared to CR+DES. This suggests that algorithm selection should be guided by the primary objective—CR+DES for compatibility-focused optimization and MSBX+MO for price-focused scenarios.

The solution generated by the Single Objective Approach (SOA) is outperformed by some specific solutions from CR+DES, CR+DES\_E, Mutation+SBX NSGA-II, and Mutation+SBX NSGA-II\_A only. These findings indicate that SOA may be a promising approach for land-use optimization. Furthermore, it is important to note that for SOA, the better performing coefficient value determined through experimentation closely aligns with the coefficient value predicted through analysis. This suggests that the coefficient values utilized within the SOA methodology may be predicted almost accurately. But, still, as expected, SOA is unable to beat the best MO methods. A notable advantage of SOA is that it is the fastest method (Supplementary Table S7).

Depending on the choice of the optimization algorithm, the land-use distribution for the area or plot count varies. The reason behind this may be that different algorithms were looking at different places of the search space, and also, there may be multiple local optima. Therefore, it might be wise to employ a set of well-performing algorithms (e.g., CR+DES and Mutation+SBX NSGA-II) followed by a (visual) comparison of the (combined) Pareto Front by the domain experts before the final selection. In this connection, an interesting idea is to employ a supervised learning component to choose one (or a few to further aid the domain expert in choosing the final one) from among the solutions provided by this set of well-performing algorithms. Such a learning based scheme to select one or a small set of promising solutions from among the set of (Pareto) solutions has recently been proposed with some success in a study \cite{kabir2023multi}, albeit in a different domain.

Finally, a brief discussion on the performance (or lack thereof, to be specific) of Two Archive based algorithm is in order. The Two Archive based algorithm, even at the 150th iteration, had individuals very close to those created while initializing. This may be attributed to the fact that convergence (finding the optimal solution) and diversity (exploring a wide range of solutions) within the two separate archives may have been unbalanced.

Our study has three main limitations. First, we only applied constraint relaxation to algorithms that already performed best—less effective methods might also improve and deserve evaluation. Second, we kept critical plots (educational institutes and hospitals) fixed; permitting controlled changes there could broaden the search space while still safeguarding essential facilities. Third, our experiments were confined to a pre-planned residential area; extending to less-structured zones like Razarbagh in Dhaka would test robustness. For future work, we will evaluate lower-ranked algorithms under relaxation, introduce flexibility in key plots, and expand our geographic scope. We also plan to integrate additional metaheuristics—NSGA-III, ant and bee colony optimization, and simulated annealing—to enrich land-use planning in Bangladesh, curb haphazard development, and reduce disaster risk. We view this study as a stepping stone toward more comprehensive, resilient urban planning strategies.

\section{Conclusion} \label{sec:n}
This paper presented a comprehensive study on computational intelligence-based land-use allocation approaches for mixed-use areas, with a particular focus on the challenging urban landscape of Dhaka, Bangladesh. Through the application of various optimization algorithms, including single objective approaches, NSGA-II variants, Two Archive algorithms, and novel custom algorithms, we have demonstrated significant improvements in both compatibility and price objectives compared to existing state-of-the-art methods.

Our key findings include: (1) The CR+DES approach emerged as the best-performing algorithm, achieving a 3.16\% improvement in compatibility compared to the state-of-the-art, while maintaining competitive price performance. (2) The introduction of relaxed constraints during the optimization process proved beneficial, allowing algorithms to explore the search space more effectively while still producing valid solutions. (3) Our custom algorithms, particularly MSBX+MO, showed superior performance in terms of hypervolume indicators and price optimization. (4) Statistical analysis using Kruskal-Wallis tests and Bonferroni-corrected Dunn's post-hoc comparisons with compact letter displays confirmed the significance of our improvements and revealed that algorithms incorporating difference vectors consistently outperform traditional approaches across multiple metrics. (5) A critical finding is that algorithm choice should be objective-dependent: CR+DES excels in compatibility-focused scenarios while MSBX+MO is superior for price optimization.

The practical implications of this research extend beyond Bangladesh to other rapidly urbanizing regions facing similar land-use allocation challenges. The methodology provides urban planners and policymakers with computational tools that can balance multiple conflicting objectives while respecting domain-specific constraints. The flexibility offered by constraint relaxation techniques enables exploration of alternative solutions that might not be immediately apparent through traditional planning approaches.

Future work will focus on extending the methodology to less planned urban areas, incorporating additional optimization algorithms such as swarm intelligence techniques, and developing supervised learning components to assist domain experts in solution selection. The ultimate goal is to contribute to more resilient, sustainable, and well-organized urban landscapes that can better serve growing urban populations while minimizing disaster risks associated with poor land-use planning.

\section*{Supplementary Materials}
The supplementary materials are available \href{https://docs.google.com/document/d/e/2PACX-1vS0eBgHcW2EpX7NWe2yGwz2ElhU-sGCNYO9wal5rXD2FmnBTiyVYs6wS3EbACYuNrQIfflqb1iTLQ_z/pub}{here.}

\section*{Declaration of Generative AI and AI-Assisted Technologies}
During the preparation of this work, the authors used ChatGPT 3.5 and Grammarly to improve grammar. After using these services, the authors reviewed and edited the content as needed and have taken full responsibility for the content of the publication.

\bibliographystyle{elsarticle-num-names}
\bibliography{citations}

\end{document}